\documentclass{fairmeta}
\usepackage{amsmath}
\usepackage{enumerate} 
\usepackage{bm}
\usepackage{floatflt}
\usepackage{algpseudocode}
\usepackage{amsfonts}
\usepackage{amsthm}
\usepackage{newtxtt}
\usepackage{colortbl} 
\usepackage{cleveref}
\usepackage{diagbox} 
\usepackage[utf8]{inputenc}
\usepackage{textgreek}
\usepackage{colortbl}
\usepackage{nicematrix}
\usepackage{makecell}
\usepackage{float}
\usepackage{arydshln}
\usepackage[frozencache,cachedir=.]{minted}

\usepackage{subcaption} 

\usepackage[most]{tcolorbox}
\usepackage{listings}
\usepackage{amssymb}
\usepackage{xspace}
\usepackage{wrapfig}
\usepackage{adjustbox}
\usepackage{tabularx}
\usepackage{booktabs}
\usepackage{mathtools}
\usepackage{wrapfig}
\usepackage{amssymb}
\usepackage{graphicx}

\usepackage[ruled,vlined]{algorithm2e} 

\lstdefinestyle{promptstyle}{
  basicstyle=\ttfamily\scriptsize,
  breaklines=true,
  breakatwhitespace=false,
  columns=fullflexible,
  keepspaces=true,
  showstringspaces=false,
  frame=none
}

\usepackage{silence}
\makeatletter
\patchcmd{\wrong@fontshape}{\@gobbletwo}{}{}{}
\makeatother
\newtheorem{theorem}{Theorem}[]

\newtheorem{remark1}[theorem]{Remark}

\definecolor{upColor}{RGB}{17,138,21}
\definecolor{downColor}{RGB}{174,36,67}

\title{VISTA: Vision-Grounded and Physics-Validated
Adaptation of UMI data for VLA Training}
\author[1,3,\dagger]{\text{Siyuan Yang}}
\author[1,4,\dagger]{\text{Linzheng Guo}}
\author[1,4]{\text{Ouyang Lu}}
\author[5]{\text{Zhaxizhuoma}}
\author[1,6]{\text{Daoran Zhang}}
\author[1,7]{\text{Xinmiao Wang}}
\author[6]{\text{Ting Xiao}}
\author[1,\ddagger]{\text{Fangzheng Yan}}
\author[1,\ddagger]{\text{Zhijun Chen}}
\author[2,8\ddagger]{\text{Yan Ding}}
\author[2,\star]{\text{Chao Yu}}
\author[1,\star]{\text{Chenjia Bai}}
\author[1,\star]{\text{Xuelong Li}}
\affiliation[1]{Institute of AI (TeleAI), China Telecom}
\affiliation[2]{Lumos Robotics}
\affiliation[3]{University of Science and Technology of China}
\affiliation[4]{Northwestern Polytechnical University}
\affiliation[5]{Shanghai Jiao Tong University}
\affiliation[6]{East China University of Science and Technology}
\affiliation[7]{Harbin Engineering University}
\affiliation[8]{Fudan University}
\contribution{\textsuperscript{$\dagger$}Equal Contribution$\quad$ 
\textsuperscript{$\ddagger$}Project Lead$\quad$
\textsuperscript{$\star$}Corresponding Authors}
\date{May, 2026}
\project{\url{https://tele-umi-vista.github.io}}
\code{\url{https://github.com/TeleHuman/umi-vista}}
\metadata[Correspondence to]{Chenjia Bai (\email{baicj@chinatelecom.cn})}

\begin{document}

\abstract{
Universal Manipulation Interface (UMI) enables scalable real-world robot data collection without hardware-specific teleoperation, yet leveraging UMI data to train large-scale Vision-Language-Action (VLA) models remains fundamentally challenging. We identify two critical mismatches: wrist-mounted fisheye views, with severe radial distortion and local gripper-centric perspectives, are out-of-distribution for pretrained VLMs; and human-collected trajectories frequently violate kinematic limits, incur collisions, or exceed controller bandwidth, teaching VLA policies physically infeasible actions. To address the challenges, we present \textbf{VISTA}, a framework that bridges this dual gap through three synergistic components. (i)~UMI-VQA, the first large-scale VQA dataset tailored to wrist-mounted fisheye observations, aligns VLM representations to the distorted visual regime via auxiliary vision-language supervision. (ii)~A systematic physical-validation pipeline performs a data-completeness pre-check and scores each valid trajectory for trajectory continuity, self-collision risk, and execution fidelity before it enters training. (iii)~A two-stage co-training recipe jointly learns vision-language grounding on UMI-VQA and action prediction on validated trajectories. Our experiments empirically show that incorporating UMI-VQA consistently improves downstream policy performance, and that physical-validation scores are strongly predictive of deployment success. On diverse simulation and real-world manipulation tasks, VISTA significantly outperforms strong baselines including $\pi_{0.5}$, LingBot-VLA, and Wall-X. We release the physical-validation pipeline, UMI-VQA, validated trajectory data, and the pre-trained model for the community.}


\maketitle

\section{Introduction}
\label{sec:intro}

Universal Manipulation Interface (UMI)~\citep{umi2024} and its successor FastUMI~\citep{Fastumi} have demonstrated that handheld gripper interfaces offer a scalable pathway to real-world robotic data collection. By equipping a human-operated gripper with a wrist-mounted fisheye camera and onboard tracking sensors, these systems capture first-person visual observations together with explicit end-effector trajectories and gripper states, all without additional data collection to a specific robot platform. The resulting datasets have proven highly effective for training compact imitation-learning policies such as Diffusion Policy~\citep{DP} and ACT~\citep{ACT}, enabling impressive real-world deployment on various tasks. However, leveraging UMI-collected data to train large-scale Vision-Language-Action (VLA) models---such as OpenVLA~\citep{Openvla}, and $\pi$-series~\citep{pi0,pi05}---presents a qualitatively different challenge. VLA models rely on powerful Vision-Language Model (VLM) backbones pretrained on massive internet-scale corpora, and they derive their generalization from deep cross-modal alignment among vision, language, and low-level action. When UMI data is used for VLA training, we observe limited gains and unreliable real-world deployment, not because the data is intrinsically unsuitable, but because the observation and execution assumptions of UMI differ fundamentally from those assumed by VLA pretraining and deployment. 

We summarize the  manifests along two critical and largely orthogonal axes: visual grounding and physical plausibility. 
(i) For \textbf{visual grounding}, contemporary VLA models are typically trained with robot demonstrations and auxiliary vision-language supervision that provide relatively global scene context. Many robot datasets include third-person or main-view observations, and action-free vision-language data used for co-training is often collected from standard perspective images. In these data sources, projection geometry is relatively regular, scene layouts are stable, and spatial cues are often globally visible. In contrast, UMI and FastUMI collect demonstrations from wrist-mounted fisheye cameras attached to the handheld gripper, e.g., cameras with a 180$^\circ$ field of view. The resulting observations are local, gripper-centric, and substantially different from global or main-view visual supervision. In addition to this viewpoint shift, fisheye projection introduces severe radial distortion and highly non-uniform spatial resolution: the image center preserves fine detail while peripheral regions are heavily compressed. Moreover, wrist-mounted placement introduces frequent self-occlusion from the gripper or robot arm.
Together, the geometric warping induced by fisheye projection and the wrist-only, gripper-centric viewpoint shift UMI observations away from the visual distributions commonly seen during VLM pretraining and VLA co-training, rendering them effectively out-of-distribution for existing visual representations;
(ii) For \textbf{physical plausibility}, the very freedom that makes UMI scalable---humans can demonstrate anywhere without robot hardware---also severs the link between collected trajectories and the physical constraints of downstream target embodiments. First, regarding \emph{kinematic constraints}, FastUMI records handheld gripper poses via onboard tracking modules (e.g., RealSense T265). Yet during collection, the recorded trajectories are not constrained by the target robot’s joint limits, reachable workspace, or motion-speed limits. They may therefore contain kinematically unreachable poses, tracking-induced discontinuities or abrupt jumps, or motions that require excessively high joint velocities during inverse-kinematics replay. Second, \emph{collision constraints} are entirely absent during collection. The tracking system monitors only the gripper pose, not the full robot body; the human operator naturally avoids environmental obstacles by repositioning the gripper, but the robot's elbow, torso, or base may still collide with each other during deployment. Third, \emph{tracking and execution constraints} are not enforced during collection. Recorded trajectories may require motions that exceed the target robot controller's bandwidth or tracking capability, and such infeasibilities are often exposed only during replay or deployment. When a VLA model learns from trajectories that violate kinematic limits, incur self-collisions, or exceed controller tracking bandwidth, it internalizes not only manipulation skills but also \emph{physically hallucinated} action patterns that cause systematic deployment failures.

To bridge this dual gap, we introduce \textbf{VISTA}, a VLA adaptation framework that aligns UMI-collected data with the perceptual and physical requirements of generalist robot policies through three synergistic components. (i) To resolve the visual grounding mismatch, we construct \textbf{UMI-VQA}, which is, to our knowledge, the \emph{first} large-scale vision-language dataset tailored to wrist-mounted fisheye observations. UMI-VQA contains 8M question-answer pairs grounded in the same fisheye visual regime, covering scene understanding, interaction grounding, and spatial reasoning. By co-training the VLM backbone on UMI-VQA alongside action data, we align its visual representations to the distorted geometry and local first-person perspective inherent in wrist-mounted fisheye views, rather than forcing costly fine-tuning from scratch. (ii) To ensure physical plausibility, we propose a systematic \textbf{physical validation} pipeline applied to every trajectory before it enters the VLA training. Unlike coarse per-pose filtering, our validation audits entire trajectories along three dimensions: trajectory-level kinematic reachability and smoothness, self-collision checks, and controller tracking-feasibility analysis. Only trajectories that pass three audits are retained for training, thereby guaranteeing that the VLA model learns exclusively from embodiment-compatible and physically consistent demonstrations. (iii)~We train VISTA with a two-stage co-training recipe. Stage one performs joint autoregressive learning on large-scale UMI-VQA and validated UMI discrete action token to establish aligned vision-language-action representations. Stage two refines continuous control generation via a flow-matching action expert~\citep{lipman2023flow,ReFlow}, which captures the multimodal, high-dimensional action distributions. We pre-train the model on 100K real-world UMI trajectories that pass our physical validation, together with 8M UMI-VQA samples, to obtain the final pre-trained VISTA model. 

We evaluate VISTA through three tiers of experiments designed to isolate and validate each design choice. \textbf{(1)~Diagnostic validation:} We show that current state-of-the-art embodied-specific VLMs suffer significant degradation in visual understanding and spatial reasoning when evaluated on fisheye-adapted benchmarks; we further demonstrate that a large fraction of raw UMI trajectories cannot be replayed on real robots due to kinematic limits, collisions, or tracking errors, confirming the empirical validity of our stated challenges. 
\textbf{(2)~Data-level validation:} We verify that co-training with UMI-VQA improves downstream policy performance compared with action-only training and standard-view VQA supervision. Moreover, through score-controlled subset experiments, we find that higher physical-validation scores generally correspond to better deployment outcomes, suggesting that our validation metric is a useful proxy for data utility. 
\textbf{(3)~Model-level validation:} We fine-tune and evaluate VISTA on UMI-style simulation benchmarks, including RoboTwin-UMI and LIBERO-UMI, as well as 20 diverse real-world manipulation tasks. Under controlled same-data conditions, VISTA outperforms strong baselines including $\pi_{0.5}$, LingBot-VLA, and Wall-X. Our contributions are summarized as follows:
\begin{itemize}
    \item We identify and formalize two critical bottlenecks in adapting VLA models to UMI data: a \textbf{Visual Grounding mismatch} between wrist-mounted fisheye observations and standard-perspective VLM pretraining domains, and a \textbf{Physical-Plausibility mismatch} between human-collected trajectories and target-robot embodiment constraints.
    
    \item We propose \textbf{VISTA}, a UMI-oriented VLA framework comprising UMI-VQA for perceptual alignment, a systematic trajectory-level physical validation pipeline for embodiment-aware data curation, and a two-stage co-training recipe with a flow-matching action expert.
    
    \item We introduce and release \textbf{UMI-VQA}, the first large-scale VQA dataset for wrist-mounted fisheye observations (8M samples), and release physically validated UMI trajectories to support future research in scalable robot learning.
    
    \item We demonstrate on 20 real-world tasks and two UMI-style simulation benchmarks that VISTA substantially exceeds the performance of existing VLA baselines, establishing that explicit visual-grounding and physical-plausibility alignment are essential for unlocking the value of handheld demonstration data in generalist robot policy learning.
\end{itemize}

\section{Related Work}

\textbf{Scalable Robot Data Collection.} Large-scale robot learning hinges on diverse, action-labeled datasets. 
Conventional approaches collect demonstrations via teleoperation on physical robot platforms~\citep{Robonet, Droid, Bridgedatav2, Rh20t, Open-x,robocoin}, which provides precise action supervision but remains costly, labor-intensive, and tightly coupled to specific hardware stacks. 
Because such data reflects the particular sensor configurations, kinematics, and actuators of the collection platform, policies trained on one embodiment rarely generalize to others without extensive domain adaptation. 
To break this hardware dependence, Universal Manipulation Interface (UMI)~\citep{umi2024} introduces a handheld paradigm in which human operators freely manipulate a portable gripper equipped with a wrist-mounted fisheye camera, recording wrist-mounted fisheye cameras and end-effector trajectories without requiring a physical robot at collection time. 
Follow-up works like FastUMI \citep{Fastumi,Fastumi100k}, DexUMI~\citep{Dexumi}, ActiveUMI~\citep{Activeumi}, UMI-3D~\citep{UMI-3D}, and RDT2~\citep{RDT2} refine tracking accuracy, expand dataset scale, and broaden gripper compatibility, demonstrating encouraging cross-embodiment results for compact imitation-learning policies without requiring explicit model-side embodiment adaptation, such as aligning target action distributions under embodiment shifts~\citep{ATE}. Yet these efforts largely treat UMI data as directly consumable by any downstream learner. We contend that this assumption collapses when the downstream model is a large-scale VLA system: the wrist-mounted fisheye visual distribution and the physically unconstrained nature of human-collected trajectories introduce fundamental mismatches---both perceptual and physical---against the pretraining and deployment assumptions of modern VLM architectures~\citep{paligemma,paligemma2,gr00tn1}. Our work identifies and closes these gaps to make UMI data truly effective for VLA training. 

\textbf{VLA Models and Visual Grounding.} 
Modern VLA models~\citep{RT-1, Rt-2, Openvla, pi0, pi05, pi07, rynnvla, prts} build upon large-scale Vision-Language Models (VLMs)~\citep{paligemma2,Qwen2-vl,Qwen3-vl} that provide strong pre-trained capacities for visual feature extraction, spatial reasoning, and language grounding. 
By conditioning low-level robot actions on these rich visual-linguistic representations, VLA systems achieve impressive generalization across language-conditioned manipulation tasks~\citep{robotwin,roboarena,maniparena}. When robot observations stem from fixed main-view cameras, or conventional egocentric frames, they align well with the visual distribution of VLM pretraining corpora, enabling effective knowledge transfer. 
In contrast, UMI data is captured from wrist-mounted fisheye cameras with extreme radial distortion, highly non-uniform spatial resolution, and severe self-occlusion by the gripper and arm~\citep{Fastumi,umi2024}. 
This creates a severe domain gap: the distorted, local, gripper-centric visual regime diverges sharply from the pinhole-perspective, globally structured images that dominate VLM pretraining. 
FastUMI~\citep{Fastumi} attempts to mitigate this gap by post-processing fisheye frames with monocular depth estimators to produce pseudo-depth maps, but this merely augments a visually novel modality rather than aligning the backbone's visual grounding to the fisheye domain.


\textbf{Physical Validation of UMI Demonstrations.}
Beyond perception, UMI-style learning must also account for whether collected demonstrations are physically executable on the target robot embodiment. To prevent physically infeasible action patterns from corrupting the training process, prior UMI-style systems~\citep{umi2024,xrzreo-g0} adopt a hard-filtering approach that simply discards trajectories violating joint limits or dynamic feasibility. In contrast, VISTA adopts a soft-validation approach: it continuously scores entire trajectories on continuity, self-collision risk, and execution fidelity, allowing for more nuanced selection of UMI demonstrations based on target-robot executability. Together with UMI-VQA data construction for visual grounding, such trajectory-level validation allows VISTA to address both perceptual alignment and physical plausibility in UMI-style learning.

\section{Method}
\label{sec:method}

We study language-conditioned manipulation policies learned from UMI-style demonstrations. At each timestep $t$, the policy $\pi_\theta$, parameterized by $\theta$, receives a natural-language instruction $l$, a visual observation $o_t$ comprising paired left- and right-wrist-mounted fisheye camera views, and a proprioceptive robot state $s_t$. The policy predicts an action chunk of horizon $H$:
\begin{equation}
    a_{t:t+H-1} \sim \pi_\theta(\cdot \mid o_t, s_t, l),
\end{equation}
where $s_t$ denotes the proprioceptive robot state, such as the current end-effector pose and gripper width, and $a_{t:t+H-1} = \{a_t, a_{t+1}, \ldots, a_{t+H-1}\}$ denotes a sequence of $H$ future robot actions. The objective is to learn $\pi_\theta$ from large-scale UMI corpora such that it generalizes to diverse language-conditioned tasks while remaining physically executable on the target robot embodiment.

Standard VLA pretraining assumes that visual observations align with the VLM's pretraining domain and that action labels reflect physically feasible robot motions. Raw UMI data satisfies neither assumption. Perceptually, the observations $o_t$ exhibit severe radial distortion, non-uniform spatial resolution, and gripper-centric local perspective, placing them out-of-distribution for VLMs trained on standard-perspective images. Physically, the trajectories are collected by humans who have no awareness of the target robot's joint limits, collision geometry, or controller bandwidth; they therefore serve as unreliable direct supervision for a VLA policy. Consequently, raw UMI demonstrations cannot be fed into a VLA training pipeline without first resolving both the visual grounding and physical plausibility gaps.

To bridge this dual gap, we introduce \textbf{VISTA}, a framework that converts raw UMI data into aligned, validated, and VLA-compatible training corpora. In the following, we start by introducing the UMI hardware, followed by three synergistic stages. (i)~\emph{Perception alignment} via UMI-VQA: we construct the first large-scale vision-language dataset tailored to wrist-mounted fisheye observations, and use it to adapt the VLM backbone to the distorted, gripper-centric visual regime during joint VQA-action co-training. (ii) \emph{Physical validation}: every trajectory is audited against the target robot's kinematics, collision geometry, and controller characteristics; only trajectories that pass these checks are retained for training. (iii) \emph{Two-stage co-training}: we first perform autoregressive vision-language-action co-training on UMI-VQA and validated UMI trajectories to align cross-modal representations, then refine continuous action generation with a flow-matching action expert. The resulting VISTA model can be directly fine-tuned on downstream embodiment-specific validated data for real-world deployment.

\subsection{UMI Hardware for Cross-embodiment Data Collection}
\label{sec:lumos-umi}

\begin{figure}[t]
    \centering
    \includegraphics[width=0.9\textwidth]{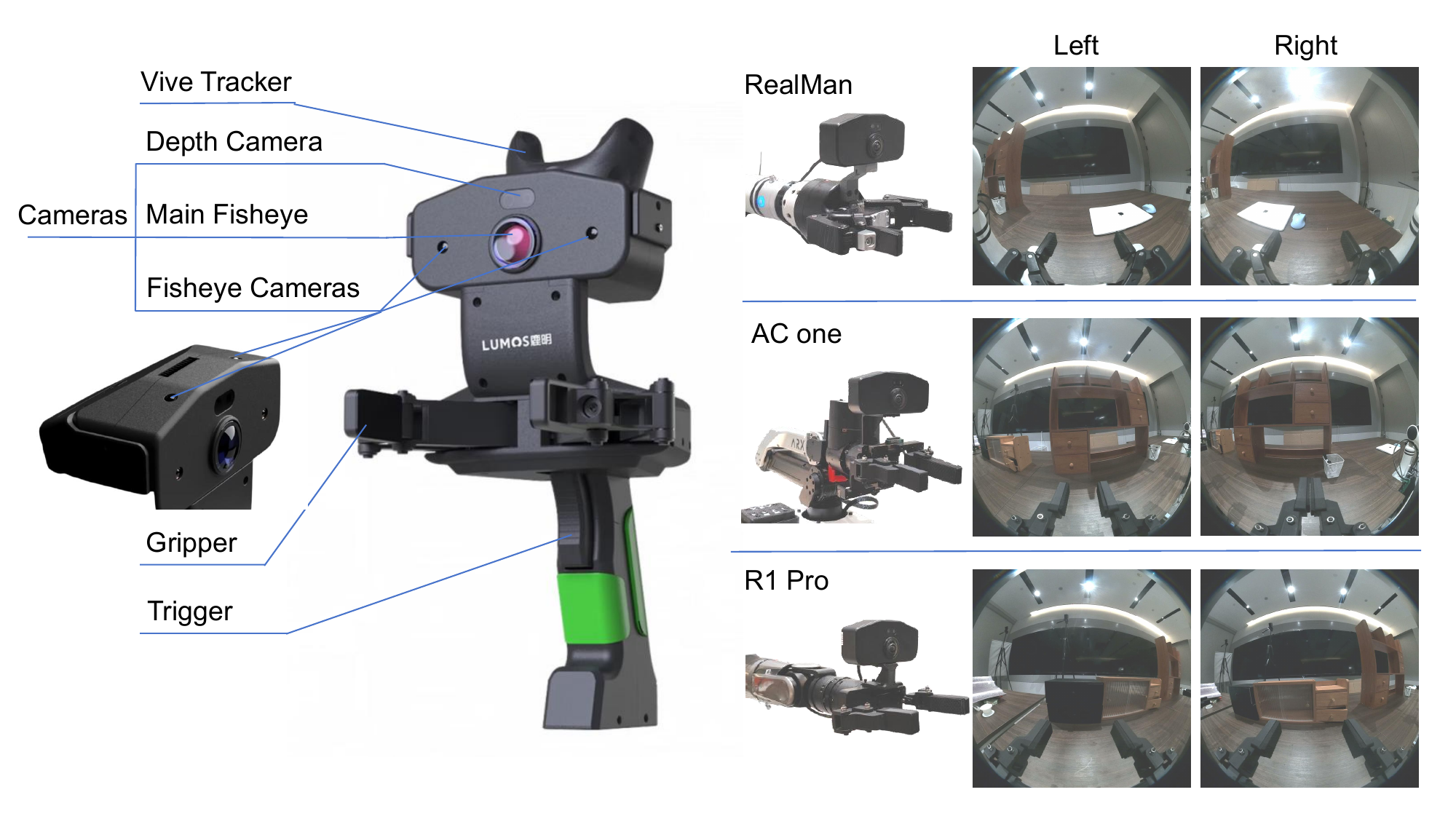}
    \caption{FastUMI Pro and observation examples. 
    (\emph{Left}) The handheld data-collection device, with labeled sensors: a central main fisheye camera, two lateral fisheye cameras, a depth camera, a Vive Tracker for 6-DoF pose estimation, and a trigger-actuated gripper with encoder-based width sensing. 
    (\emph{Right}) The same wrist-mounted camera configuration is deployed on three dual-arm robot platforms (RealMan, AC one, and Galaxea R1 Pro). For each platform we show the paired left- and right-wrist fisheye observations; note the purely gripper-centric viewpoint with no external main-view camera, enabling portable in-the-wild collection but also introducing severe radial distortion and self-occlusion.}
    \label{fig:umi-hardware}
\end{figure}

We use the FastUMI Pro system from \emph{Lumos Robotics} as our data-collection platform (Fig.~\ref{fig:umi-hardware}, left). The handheld gripper integrates a multi-sensor suite in a lightweight, self-contained form factor ($\sim$600~g).
A main fisheye camera is mounted centrally above the gripper jaws, providing a wide-angle ($\approx$180$^\circ$ diagonal field of view) RGB observation of the local workspace. Flanking the main camera are two auxiliary fisheye cameras that together with an onboard depth camera constitute a quad-ocular visual system; this redundancy maintains robust visual tracking even when the central view is occluded by the gripper fingers or suffers from texture-poor regions. End-effector pose is recovered by fusing two independent tracking streams: a Vive Tracker that provides 6-DoF pose via an external lighthouse system, and a real-time visual-inertial SLAM pipeline running on the onboard camera array.
The fused pose estimate achieves sub-centimeter accuracy ($\sim$3~mm) and is robust to transient occlusion, which is critical for maintaining data quality during contact-rich manipulation. Gripper width is measured by an internal encoder and is actuated by a handheld trigger, allowing the human operator to control aperture continuously during demonstration. 

A key design feature of this platform is its local, gripper-centric sensing: the observations $o_t$ comprise paired left- and right-wrist-mounted fisheye views from the two gripper jaws, with no external main-view or third-person camera. This enables flexible data collection in both indoor and outdoor environments, but it also means the visual input is subject to severe radial distortion, non-uniform resolution, and extreme self-occlusion by the gripper and operator hands---properties that are central to the visual-grounding challenge. For downstream policy execution we mount the same wrist-camera configuration and end-effectors on distinct dual-arm embodiments: RealMan, AC one, and Galaxea R1 Pro (Fig.~\ref{fig:umi-hardware}, right). 
The end-effector cameras are identical to those on the handheld device, ensuring that the visual observation distribution at deployment matches the training distribution. The gripper jaws on the robot side are mechanically adapted to each platform while preserving the same fingertip geometry and camera extrinsics, so that policies trained on handheld data can be transferred without recalibrating the visual frame.

\begin{figure}[tbp]
    \centering
    \includegraphics[width=1\textwidth]{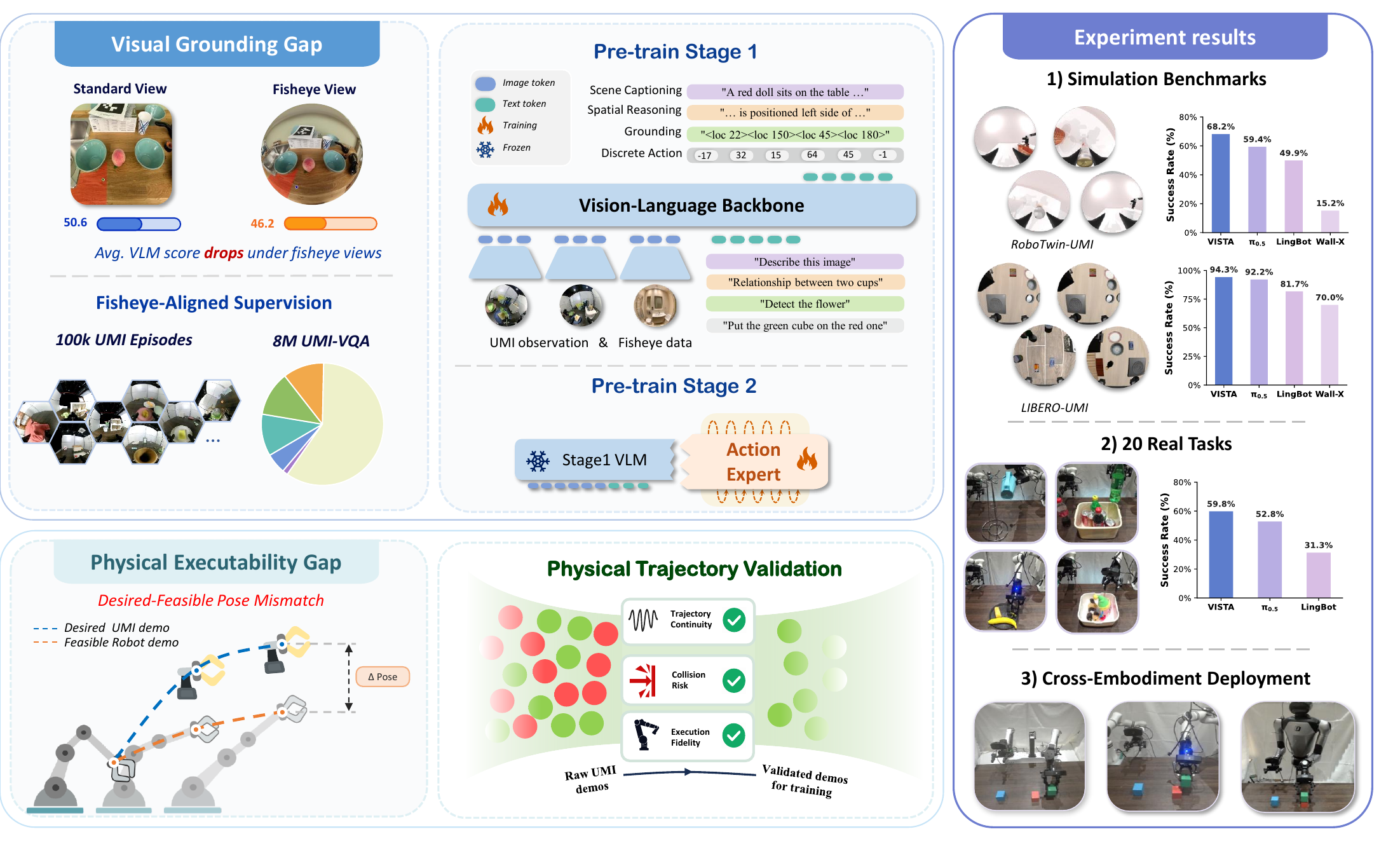}
\caption{
Raw UMI data poses two critical mismatches for VLA training: wrist-mounted fisheye views are out-of-distribution for pretrained VLMs, and human-collected trajectories may be physically infeasible for target robot embodiments. 
VISTA addresses these challenges with an 8M-sample UMI-VQA dataset for fisheye vision-language alignment, a physical-validation pipeline for kinematic reachability, collision freedom, and tracking feasibility, and a two-stage VQA-action co-training recipe. 
Across UMI-style simulation, real-world, and cross-embodiment experiments, VISTA significantly outperforms strong baselines.
}
\label{fig:example}
\end{figure}

\begin{figure}[h!]
    \centering
    \includegraphics[width=1\textwidth]{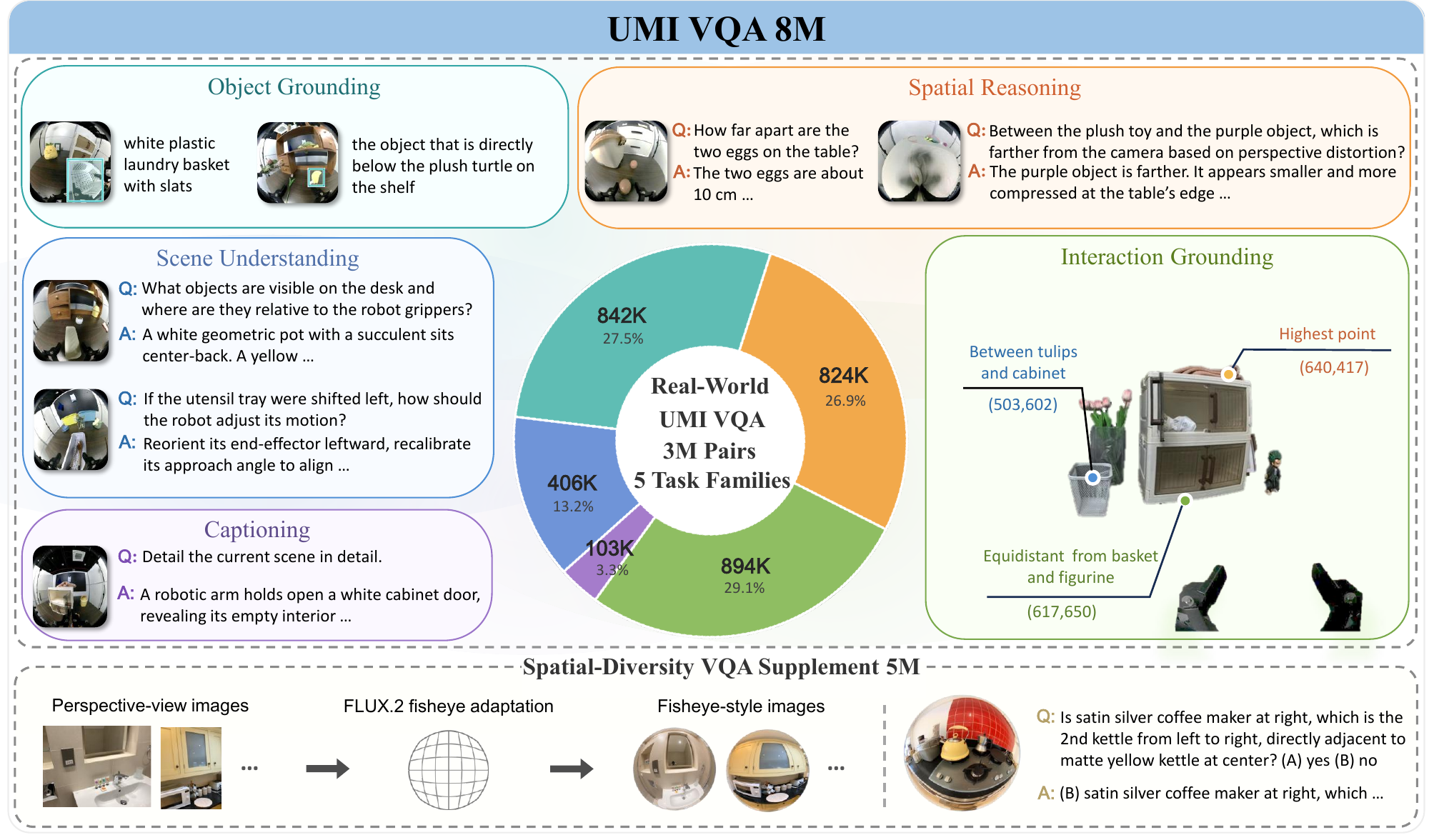}
    \caption{Overview of UMI-VQA. UMI-VQA contains 8M samples from two complementary sources: 3M real-world wrist-fisheye UMI VQA pairs organized into five capability-oriented subsets—Object Grounding (842K, 27.5\%), Scene Understanding (406K, 13.2\%), Captioning (103K, 3.3\%), Interaction Grounding (894K, 29.1\%), and Spatial Reasoning (824K, 26.9\%)—and a 5M spatial-diversity VQA supplement adapted from perspective-view images into fisheye-style images.}
    \label{fig:example-umi-vqa}
\end{figure}

\subsection{UMI-VQA for Perception Alignment}
\label{sec:umi_vqa}

To address the visual-grounding mismatch from limited UMI data, we construct \textbf{UMI-VQA}, a large-scale vision-language dataset that provides auxiliary perceptual supervision within the same fisheye observation regime as the action data. UMI-VQA is built from two complementary visual sources that together cover both authentic fisheye geometry and diverse scene layouts. 
\begin{itemize}
    \item \textbf{Real-world UMI demonstration frames.} The dominant portion of UMI-VQA is derived from real-world UMI-style manipulation videos. We sample frames from trajectories collected by the handheld device described in Sec.~\ref{sec:lumos-umi}, preserving the full suite of visual properties encountered during policy learning: severe radial distortion, non-uniform spatial resolution, extreme self-occlusion by the gripper and operator hands, and local gripper-centric viewpoints. These frames are annotated through a semi-automatic pipeline: a large-scale VLM \citep{Qwen3-vl} is prompted to generate structured descriptions covering scene content, object states, and spatial relations, followed by lightweight human verification to ensure factual correctness and manipulation relevance. Because these annotations are grounded in authentic fisheye observations, they directly adapt the VLM to the exact visual distribution that the policy will encounter at deployment. 
    \item \textbf{Edited spatial-diversity images.} While real UMI trajectories provide fisheye geometry, they remain limited in environmental diversity due to the constrained scope of our data collection, which covers a relatively narrow range of scenes, layouts, and object configurations. To broaden coverage of spatial relations and scene configurations, we supplement the real frames with images from RefSpatial~\citep{roborefer}, a large-scale dataset emphasizing 3D spatial reasoning. However, these images are captured from standard perspectives and must be brought into the fisheye domain. A naive approach would be to apply classical geometric warping (e.g., polynomial distortion) to remap pinhole-perspective images into fisheye projections. However, we find it is insufficient as conventional perspective-to-fisheye warping is a pure geometric pixel-remapping operation; when the target fisheye field of view exceeds the original image's visible scope---which is precisely the case for lenses with $\approx$180$^\circ$ diagonal FoV---such remapping inevitably produces stretched boundaries, unnatural extrapolation, or missing content in the periphery, because geometric transforms cannot synthesize plausible scene structure beyond the original viewing frustum. In contrast, we employ a diffusion-based image-editing model (i.e., FLUX.2-dev~\citep{flux2dev}) to perform semantic-aware image-to-image translation. The model not only applies geometric distortion but also hallucinates physically plausible peripheral content conditioned on global scene semantics, yielding fisheye-style views with naturally compressed edges and coherent wide-angle perspective that more closely mimic the optical characteristics of real fisheye lenses. This generative approach preserves the spatial-relation annotations of the original RefSpatial images while placing them in a visually authentic fisheye regime. An example is given in Fig.~\ref{fig:example-umi-vqa}. 
\end{itemize}


The real-world UMI-VQA portion is organized into five capability-oriented subsets designed to supervise complementary perceptual abilities required by wrist-fisheye manipulation.
Each subset follows task-specific annotation guidelines that define the expected visual evidence, response format, and manipulation-relevant focus. 
We describe the objective of each subset as follows. 
\begin{itemize}
    \item \textbf{Scene Captioning} set provides concise contextual descriptions of wrist-mounted fisheye observations. Rather than describing only salient objects, the annotations summarize the visible scene in terms of objects, gripper-object relations, and manipulation context. This subset helps preserve manipulation-relevant scene context within the visible wrist-fisheye view, even when the input is local, distorted, and partially observed. 
    \item \textbf{Scene-State Understanding} set captures the task-relevant state of a wrist-fisheye observation. The annotations require the model to infer how the scene is configured for manipulation, including the current gripper status, relevant objects, possible obstacles, and constraints that may affect the next action. This subset encourages the model to interpret the current manipulation state rather than only recognize objects. 
    \item \textbf{Object Grounding} set associates language references with task-relevant visual targets. The referred targets are localized with bounding boxes and may be described by category names, spatial relations, or task context. This subset improves reference resolution under local viewpoints, occlusion, and fisheye distortion. 
    \item \textbf{Interaction Grounding} set identifies where manipulation should occur. Instead of only localizing entire objects, the annotations point to actionable regions such as grasp sites, contact regions, functional parts, and relation-dependent interaction points. This subset encourages the model to reason about affordances and contact geometry under gripper-centric observations. 
    \item \textbf{Spatial Reasoning} set focuses on geometric and relational understanding under fisheye observations. The annotations require the model to reason about object layout, relative position, depth, orientation, reachability, and potential collision constraints in the workspace. This subset strengthens the model's ability to interpret spatial structure from distorted wrist-mounted views.
\end{itemize}

Together, these real-world UMI VQA subsets provide structured wrist-fisheye perception supervision for semantic understanding, spatial grounding, and task-aware reasoning. Combined with the RefSpatial-based spatial-diversity supplement, UMI-VQA further expands the range of spatial relations and scene layouts covered by this perception supervision. By co-training the VLM backbone on UMI-VQA, we align its visual representations to the distorted geometry, non-uniform resolution, and gripper-centric perspective inherent in wrist-mounted manipulation, establishing a stronger perceptual foundation for downstream action learning. Fig.~\ref{fig:example-umi-vqa} summarizes the dataset structure and representative examples. The complete annotation prompts and data construction pipeline are provided in Appendix~\ref{app:umi_vqa_construction}.

\subsection{A Trajectory Scoring Framework for Physical Validation}

\begin{figure}[t]
    \centering
    \includegraphics[width=1\textwidth,trim=0bp 0bp 0bp 4bp, clip]{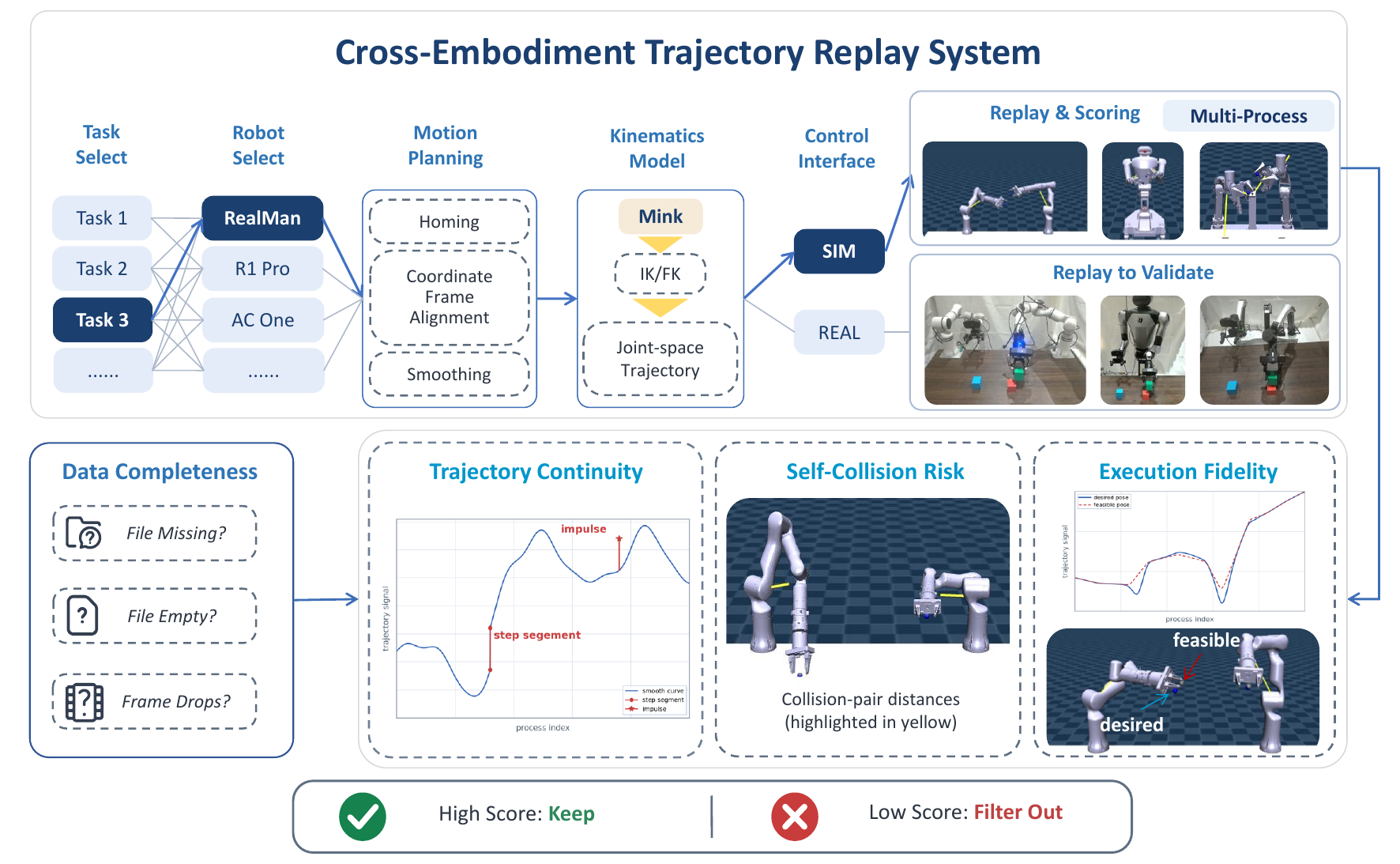}
    \caption{Cross-embodiment physical-validation pipeline. Raw UMI trajectories are replayed on target robot kinematics in simulation. Each trajectory first undergoes a data-completeness pre-check and is then scored along three axes: trajectory continuity, self-collision risk, and execution fidelity. An overall score $S(\xi,e)$ determines the evaluation score.}
    \label{fig:physical-validation}
\end{figure}

Raw UMI trajectories are collected by human operators who have no knowledge of the downstream robot embodiment. To prevent unreliable or physically infeasible demonstrations from entering the training corpus, we introduce a trajectory-level validation mechanism. It first applies a data-completeness pre-check and then scores each valid trajectory along three dimensions: trajectory continuity, self-collision risk, and execution fidelity. Here, trajectory continuity measures embodiment-agnostic recording quality and motion smoothness, whereas self-collision risk and execution fidelity evaluate embodiment-conditioned physical feasibility. Unlike binary accept/reject filters, our scoring-based formulation provides continuous, threshold-adjustable quality control that naturally supports cross-embodiment policy learning. Because the target deployment robot is unknown during pre-training, we can apply a loose validation threshold that retains trajectories with moderate scores across a diverse set of candidate embodiments. This preserves data diversity and encourages the policy to learn embodiment-agnostic action priors. During downstream fine-tuning, the threshold is tightened for the \emph{specific} target robot, ensuring that only trajectories with high kinematic compatibility and collision safety are used for specialization. This two-stage curation balances generalization and deployability. Fig.~\ref{fig:physical-validation} illustrates the physical validation process. 

A cross-embodiment trajectory replay system is developed as the basis of validation, using a MuJoCo simulation~\citep{mujoco} with kinematics computed via the Mink library~\citep{Mink}. The kinematic models of all robots are built using Mink. Once the task and robot type are selected, the robot is first initialized to home. Then a smooth trajectory passing all waypoints is planned and executed in a coordinate frame aligned with the UMI. The joint-space trajectories are derived by inverse kinematics and sent to the simulation or the real robot via the control interface. Scores are collected during the replay.

\textbf{Trajectory Continuity} ($s_{\text{tc}}$).~
Trajectory continuity measures the intrinsic smoothness of the recorded gripper motion in task space. Large discontinuities between consecutive waypoints indicate sensor dropout, tracking loss, or abrupt human motion, all of which cause unstable robot execution. This score is \emph{embodiment-agnostic} because it evaluates the raw trajectory prior to any robot-specific mapping. Given a trajectory $\xi$, let $d^p_t$ and $d^r_t$ denote the positional and angular displacements between the consecutive waypoints at timestep $t$, respectively. We first define a three-regime piecewise scoring function for a generic displacement $d$ with hyperparameters $\alpha$ and $\beta$:
\begin{equation}
g(d) = 
\begin{cases}
100, & \text{if } d \le d_{\text{min}}, \\[4pt]
100 - \alpha \dfrac{d - d_{\text{min}}}{d_{\text{max}} - d_{\text{min}}}, & \text{if } d_{\text{min}} < d \le d_{\text{max}}, \\[8pt]
\beta \exp\!\left(-\dfrac{d - d_{\text{max}}}{d_{\text{scale}}}\right), & \text{if } d > d_{\text{max}}.
\end{cases}
\end{equation}
The first regime rewards near-ideal smoothness with full marks; the second applies a linear penalty for moderate deviations; the third imposes an exponential decay for severe discontinuities that likely stem from hardware failures or tracking loss. Positional and angular components are scored separately with task-specific hyperparameters. The lower score between the positional and angular components is used as the continuity score at each timestep, and the trajectory-level score $s_{\text{tc}}(\xi)$ is obtained by taking the minimum continuity score over all consecutive waypoints. In practice, we use $d_{\text{min}}=5\,\text{mm}$, $d_{\text{max}}=45\,\text{mm}$, and $d_{\text{scale}}=100\,\text{mm}$ for translation, and $1^{\circ}$, $9^{\circ}$, and $20^{\circ}$ for rotation, with $\alpha=40$ and $\beta=60$.


\textbf{Self-collision risk} ($s_{\text{sr}}$).~
While UMI collection is embodiment-agnostic, deployment is not. We evaluate self-collision by replaying each trajectory in the cross-embodiment trajectory replay system. Given a trajectory $\xi$ and a target embodiment $e$, we record the minimum collision-pair distance $d_{\text{col}}(\xi,e)$ over all timesteps and all designated robot link pairs during replay. The self-collision score is
\begin{equation}
s_{\text{sr}}(\xi,e) = 
\begin{cases}
100, & \text{if } d_{\text{col}}(\xi,e) \ge d_{\text{col},\text{max}}, \\[4pt]
100 \cdot \dfrac{d_{\text{col}}(\xi,e) - d_{\text{col},\text{min}}}{d_{\text{col},\text{max}} - d_{\text{col},\text{min}}}, & \text{if } d_{\text{col},\text{min}} < d_{\text{col}}(\xi,e) < d_{\text{col},\text{max}}, \\[8pt]
0, & \text{if } d_{\text{col}}(\xi,e) \le d_{\text{col},\text{min}}.
\end{cases}
\end{equation}
Trajectories exhibiting link-link distances below $d_{\text{col},\text{min}}$ receive a zero score, corresponding to a hard self-collision violation, whereas those maintaining a safety margin above $d_{\text{col},\text{max}}$ are fully rewarded. Scene objects are excluded from this check to isolate self-collision from environment interaction.

\textbf{Execution fidelity} ($s_{\text{ef}}$).~
The execution fidelity measures how faithfully a target robot can reproduce the demonstrated end-effector motion. Using the same cross-embodiment replay infrastructure, we replay the UMI trajectory on the target embodiment and compute the tracking deviation between the demonstrated end-effector pose and the pose actually achieved by the robot under its joint-level controller. This deviation aggregates multiple embodiment-specific factors: proximity to joint limits, kinematic singularities, and workspace boundaries. Given a trajectory $\xi$ and a target embodiment $e$, we compute the replay tracking error $e_{\text{replay}}(\xi,e)$ over the replayed trajectory. The positional and angular replay errors are scored separately using the same functional form as $s_{\text{tc}}$, and the lower component score is used at each timestep. The minimum score over the entire replay is taken as the execution-fidelity score $s_{\text{ef}}(\xi,e)$.

\textbf{Overall cross-embodiment score.} Because $s_{\text{sr}}$ and $s_{\text{ef}}$ are inherently embodiment-dependent, the overall quality of a trajectory $\xi$ must be conditioned on a specific robot $e$. We aggregate the three scores via a weighted product model that preserves the hard-constraint nature of physical feasibility while allowing flexible emphasis:
\begin{equation}
S(\xi, e) = 100 \cdot \left( \frac{s_{\text{tc}}(\xi)}{100} \right)^{w_1} \left( \frac{s_{\text{sr}}(\xi, e)}{100} \right)^{w_2} \left( \frac{s_{\text{ef}}(\xi, e)}{100} \right)^{w_3}, \qquad w_i \ge 0,\ \sum_{i=1}^{3} w_i=3.
\end{equation}
The weights $w_i$ can be adjusted per training stage; for example, raising $w_2$ during fine-tuning prioritizes collision-free trajectories. Unless otherwise specified, we use uniform weights $w_1=w_2=w_3=1$. For pre-training across $N$ candidate embodiments, we rank trajectories by their average cross-embodiment compatibility $(S_{\text{cross}}(\xi) = \frac{1}{N}\sum_{e=1}^{N} S(\xi,e))$, ensuring that broadly feasible demonstrations contribute preferentially to the generalist policy.

\subsection{Model Training and Deployment}
\label{sec:training_recipe}

\textbf{The pre-training stages.~} 
VISTA model is initialized from the $\pi_{0.5}$ checkpoint and further adapted on large-scale perception-aligned corpora comprising 8M UMI-VQA samples and 100K real-world robot trajectories.
Because the pre-training objective is to learn embodiment-agnostic manipulation priors, the robot trajectories are curated with a lenient physical validation threshold that retains a broad spectrum of cross-embodiment behaviors while filtering only severely defective demonstrations. 
The pre-training proceeds in two stages: first, autoregressive co-training on VQA and discretized actions to align the VLM backbone with the fisheye observation regime; second, continuous-action refinement via a knowledge-isolated flow-matching expert.

\paragraph{Stage 1: VQA-Action Autoregressive Co-training.}
We first co-train the VLM backbone on action prediction and VQA answering. For action learning, each continuous action chunk $a_{t:t+H-1}$ is converted into a sequence of discrete FAST tokens $z_{1:N_a}$. Given visual observations $o_t$, language instruction $l$, and robot state $s_t$, the target output is the action-token sequence $z_{1:N_a}$. For UMI-VQA supervision, given an image observation $o$, a language question $q$, and the target answer sequence $u_{1:N_q}$, the target output is the answer-token sequence $u_{1:N_q}$. Action tokens and answer tokens are optimized with the same autoregressive next-token prediction objective. Let $D_{\text{mix}}$ denote the Stage-1 training corpus, where each example is represented as an input-output pair $(x, y)$:
\begin{equation}
(x, y) = \begin{cases}
\big((o_t, l, s_t), z_{1:N_a}\big), & \text{for action samples}, \\
\big((o, q), u_{1:N_q}\big), & \text{for UMI-VQA samples}.
\end{cases}
\end{equation}
The Stage-1 objective is:
\begin{equation}
    \mathcal{L}_{\mathrm{stage1}}(\theta)
    =
    -
    \mathbb{E}_{(x,y)\sim \mathcal{D}_{\mathrm{mix}}}
    \frac{1}{|y|}
    \sum_{j=1}^{|y|}
    \log p_\theta(y_j \mid y_{<j}, x),
\end{equation}
where $|y|$ denotes the number of target tokens in the current sample. This stage adapts the backbone to wrist-mounted fisheye observations through perception-aligned UMI-VQA supervision, while learning discrete action-token representations for subsequent continuous-control refinement.

\paragraph{Stage 2: Knowledge-Isolated Flow Matching Action Expert.} While discrete action tokens provide a convenient interface for autoregressive VLA training, continuous control benefits from a more expressive action generator.
To prevent catastrophic forgetting of the perception and discrete-action knowledge acquired in Stage 1, we follow the \emph{knowledge-isolation} strategy introduced in~\cite{KI-2025}, keeping the pretrained VLM backbone frozen and training a separate continuous action expert on top.



Specifically, given the fixed backbone representation $h_\theta(o_t, l, s_t)$, the action expert $f_\phi$ is trained with a flow-matching objective to model the conditional generation of continuous action chunks.
For a clean action chunk $a$ and noise $\epsilon$, we construct an interpolated action $a_\tau$ at time $\tau \in [0,1]$:
\begin{equation}
    a_\tau = (1-\tau)\epsilon + \tau a.
\end{equation}
The action expert predicts the target velocity field conditioned on the VLA representation:
\begin{equation}
    \mathcal{L}_{\text{fm}}
    =
    \mathbb{E}_{a,\epsilon,\tau}
    \left[
    \left\|
    f_\phi(a_\tau, \tau, h_\theta(o_t,l,s_t))
    -
    (a-\epsilon)
    \right\|_2^2
    \right].
\end{equation}
The resulting expert learns a continuous action distribution while reusing the frozen perception and language representations from Stage 1.

\begin{figure}[t]
    \centering
    \includegraphics[
        width=1.0\textwidth,
        trim=0 170 0 20,
        clip
    ]{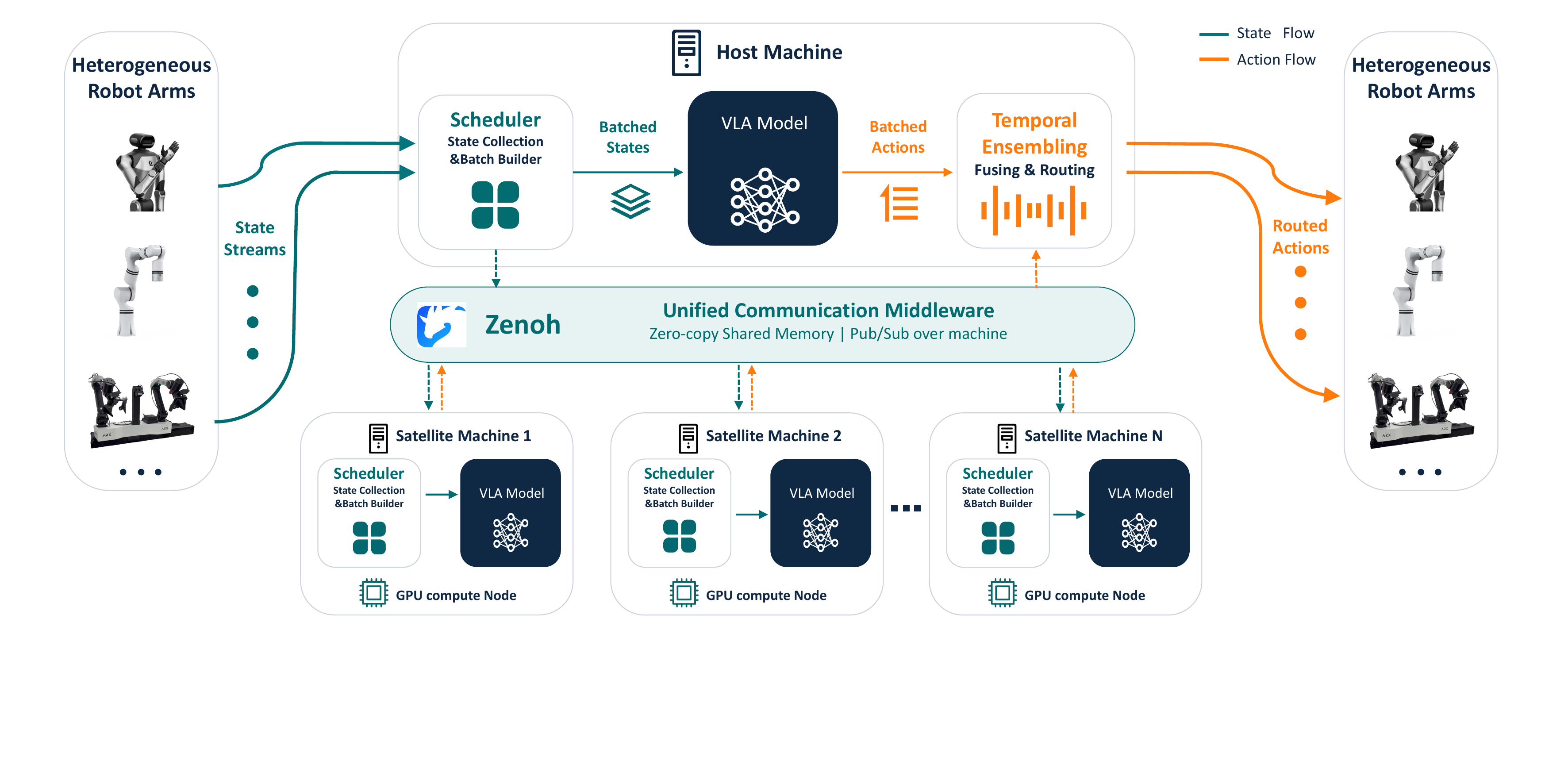}
    \vspace{-0.5em}
    \caption{The architecture of multi-machine to multi-heterogeneous robot deployment system. Heterogeneous robotic arms publish observations to Zenoh topics. Host and satellite nodes retrieve these observations to perform distributed inference. The host node subsequently handles action ensembling and routing, dispatching the final commands back to the respective robotic arms via the Zenoh middleware for execution.}
    \label{fig:multi2multi}
\end{figure}

\textbf{Downstream task fine-tuning.}
After pre-training, VISTA can be adapted to downstream tasks and target embodiments. We first apply a strict physical validation threshold to filter the downstream task data for the specific deployment robot, removing trajectories with kinematic violations, self-collision risks, or poor replay fidelity.
During fine-tuning, we unfreeze the full model and update both the VLM backbone and the flow-matching action expert end-to-end.     This adaptation protocol allows the model to preserve the generalist visual-linguistic knowledge learned during pre-training while jointly specializing perception and continuous action generation to the target robot's dynamics and task distribution.

\textbf{Multi-robot deployment system.}
To fully exploit the cross-embodiment potential of VISTA, we implement a pure Python distributed deployment architecture for heterogeneous robotic arms.
The system uses Zenoh as the communication middleware, enabling transparent shared-memory or network-level transmission across local processes, LANs, or WANs.
State streams from multiple arms are aggregated to distributed GPU compute nodes for batched inference; predicted action chunks are temporally ensembled and routed back to the respective robots via synchronous RPC calls to ensure strict temporal alignment and prevent command accumulation.
This design eliminates heavy ROS dependencies and allows seamless integration of new robot arms that satisfy the UMI end-effector mounting specification. An illustration of this system is given in Fig.~\ref{fig:multi2multi}. Further architectural details are provided in Appendix~\ref{appendix:multi-deployment}.


\section{Experiments}
\label{sec:experiments}


Our experiments are organized into three tiers that isolate and validate each design choice of VISTA.
First, \emph{diagnostic validation} confirms the empirical validity of the two core challenges: we show that state-of-the-art VLMs and VLA policies suffer significant degradation under wrist-mounted fisheye observations, and that a substantial fraction of raw UMI trajectories cannot be faithfully replayed on target embodiments due to kinematic infeasibility, collision risks, or poor replay fidelity. Second, \emph{data-level validation} verifies the efficacy of our proposed data-level remedies: we demonstrate that co-training UMI-VQA with action data consistently improves downstream policy performance over generic VQA supervision, and that physical-validation scores are strongly predictive of real-world deployment success, establishing the score as an effective proxy for data utility. Third, \emph{model-level validation} evaluates the complete VISTA system against strong VLA baselines on UMI-style simulation benchmarks and 20 diverse real-world manipulation tasks, accompanied by ablation studies on key architectural components.


\begin{table}[t]
\centering
\small
\begin{tabular}{lccccc}
\toprule
\textbf{Model}
& \multicolumn{2}{c}{\textbf{LIBERO}}
& \multicolumn{2}{c}{\textbf{RoboTwin}}
& \textbf{Avg. Drop} \\
\cmidrule(lr){2-3}
\cmidrule(lr){4-5}
& \textbf{Standard View}
& \textbf{Wrist-Fisheye}
& \textbf{Standard View}
& \textbf{Wrist-Fisheye}
&  \\
\midrule
$\pi_{0.5}$ & 96.3 & 92.2 & 82.0 & 59.4 & 13.4 \\
Wall-X      & 74.6 & 70.0 & 14.9 & 15.2 & 2.2  \\
LingBot-VLA     & 85.3 & 81.7 & 77.6 & 49.9 & 15.7 \\
\bottomrule
\end{tabular}
\caption{
Policy degradation under UMI-style wrist-fisheye observations on LIBERO and RoboTwin.
For each benchmark, we compare the separately fine-tuned standard-view policy with the wrist-only fisheye policy.
Avg. Drop reports the average performance decrease across the two benchmarks caused by the UMI-style observation regime.
}
\label{tab:policy_degradation_fisheye}
\end{table}

\begin{table*}[t]
\centering
\small
\setlength{\tabcolsep}{4.5pt}
\begin{tabular}{lccccc}
\toprule
\textbf{Model} 
& \textbf{Where2Place} 
& \textbf{RefSpatial} 
& \textbf{ERQA} 
& \textbf{EmbSpatial}
& \textbf{Avg.} \\
\midrule
Qwen2.5VL-3B        
& 0.300 / 0.200 
& 0.330 / 0.262 
& 0.328 / 0.345 
& 0.415 / 0.404 
& 0.343 / 0.303 {\small\textbf{($\bm{\downarrow}$ 11.8\%)}} \\

Qwen3VL-4B          
& 0.680 / 0.650 
& 0.485 / 0.441 
& 0.383 / 0.363 
& 0.539 / 0.539 
& 0.522 / 0.498 {\small\textbf{($\bm{\downarrow}$ 4.5\%)}} \\

Embodied-R1-3B-v1 
& 0.570 / 0.460 
& 0.398 / 0.293 
& 0.348 / 0.318 
& 0.431 / 0.437 
& 0.437 / 0.377 {\small\textbf{($\bm{\downarrow}$ 13.7\%)}} \\

RoboBrain2.5-4B   
& 0.780 / 0.690 
& 0.550 / 0.535 
& 0.353 / 0.333 
& 0.542 / 0.528 
& 0.556 / 0.522 {\small\textbf{($\bm{\downarrow}$ 6.2\%)}} \\

VLASER-2B         
& 0.690 / 0.570 
& 0.420 / 0.398 
& 0.338 / 0.325 
& 0.501 / 0.480 
& 0.487 / 0.443 {\small\textbf{($\bm{\downarrow}$ 9.0\%)}} \\
\bottomrule
\end{tabular}
\caption{
Perception degradation under fisheye observations.
Each benchmark entry reports performance on original resized images / fisheye-transformed images.
The Avg. column reports the mean performance across four benchmarks for each model, with the relative degradation shown in parentheses.
}
\label{tab:fisheye_perception}
\end{table*}

\subsection{Diagnostic Validation Experiments}
\label{sec:exp-diagnostic}

Diagnostic validation aims to verify that the two UMI-to-VLA bottlenecks---perception mismatch and physical infeasibility---are empirically real and significant.
We first show that wrist-mounted fisheye observations degrade both policy learning and visual reasoning.
We then demonstrate that raw human-collected UMI trajectories are not directly executable on target robot embodiments.

\paragraph{Fisheye observations degrade policy learning.}
We first isolate how the UMI-style observation regime affects policy learning. On RoboTwin~\citep{robotwin}, replacing the standard main-view-plus-wrist observation setup with wrist-only perspective cameras causes a severe drop in $\pi_{0.5}$ success rate to 13.1\%, indicating that narrow-FOV wrist views alone provide insufficient scene context. Expanding the wrist cameras to UMI-style wide-angle fisheye observations recovers part of this lost context, but still leaves a clear gap to the standard-view setting, as shown in Table~\ref{tab:policy_degradation_fisheye}. This suggests that UMI-style wrist-fisheye observations are more informative than narrow wrist views, yet remain challenging because the policy must rely on local, gripper-centric views with radial distortion rather than globally organized main-view observations.

We further verify this observation shift across different VLA backbones and benchmarks. As shown in Table~\ref{tab:policy_degradation_fisheye}, switching from standard-view training to wrist-fisheye training reduces the average performance on LIBERO~\citep{libero} and RoboTwin for all three models, including $\pi_{0.5}$, LingBot-VLA, and Wall-X. The degradation is particularly pronounced for $\pi_{0.5}$ and LingBot-VLA, whose average success rates drop by 13.4 and 15.7 points, respectively. Overall, these results demonstrate that UMI-style wrist-fisheye observations introduce a genuine perception bottleneck for VLA policy learning.

\paragraph{Fisheye observations degrade robot-relevant visual reasoning.}
The policy degradation above suggests that VLM backbones struggle to parse fisheye-distorted scenes.
To quantify this, we evaluate general and robot-specialized VLMs, i.e., Qwen2.5VL \citep{Qwen2.5vl}, Qwen3VL \citep{Qwen3-vl}, Embodied-R1 \citep{yuan2025embodied}, RoboBrain 2.5 \citep{robobrain25}, VLASER \citep{vlaser}, on four spatial-reasoning benchmarks---Where2Place \citep{robopoint}, RefSpatial \citep{roborefer}, ERQA \citep{erqa}, and EmbSpatial \citep{embspatial}---under both standard-perspective and fisheye-transformed images, using identical questions and evaluation protocols. As shown in Table~\ref{tab:fisheye_perception}, fisheye transformation causes a consistent mean absolute drop of 4.0 points (8.6\% relative degradation) across all models, with individual model drops ranging from 4.5\% to 13.7\%. This confirms that pretrained VLMs do not automatically preserve object and spatial understanding under the distorted, local, wrist-mounted fisheye regime, directly motivating the need for perception-aligned VQA.

\begin{figure}[tbp]
    \centering
    \includegraphics[width=1\textwidth,trim=0bp 0bp 0bp 95bp]{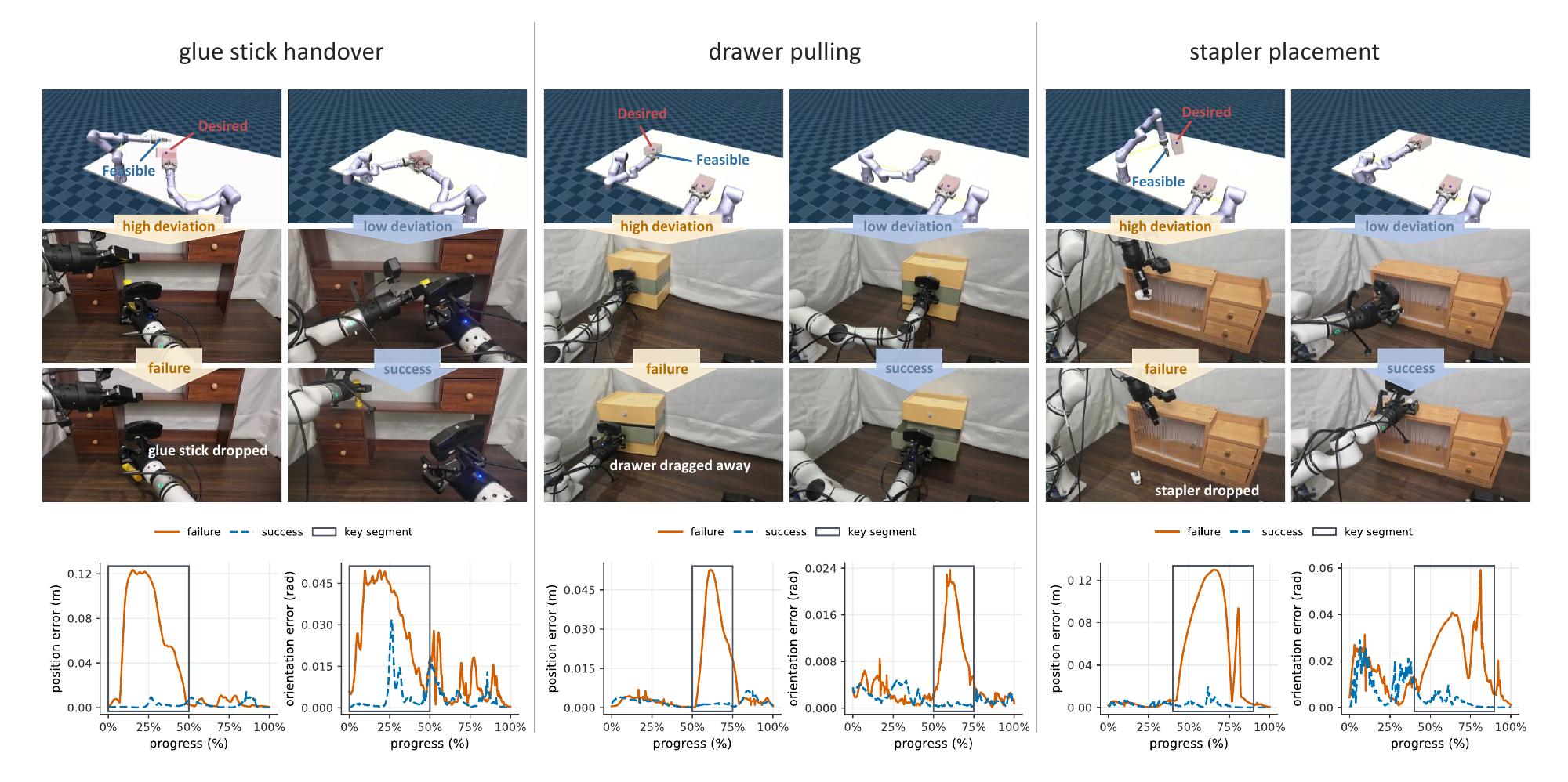}
    \caption{Comparison of UMI trajectory feasibility. The curves show the position and orientation errors computed from the discrepancy between the feasible pose and the UMI desired pose. The boxes mark the key trajectory segments where large pose deviations lead to task failures. Detailed pose trajectories are provided in Appendix~\ref{appendix:replay-details}.}
    \label{fig:feasibility_comparison}
\end{figure}

\paragraph{Raw UMI trajectories are not always executable.}
We next audit the executability of raw UMI demonstrations on three representative tasks—glue stick handover, drawer pulling, and stapler placement—performed by the RealMan robot, as illustrated in Fig.~\ref{fig:feasibility_comparison}. The raw UMI trajectories are collected and replayed both in simulation and on the physical robot. The end-effector of the robot arm is not always able to reach the desired pose, which can lead to task failure. By visually comparing the replays in MuJoCo and quantitatively analyzing the deviation between the desired and feasible poses, we observe that although the raw UMI data provides a correct trajectory, the robot's execution deviates due to its inherent constraints. This deviation is identified as the root cause of task failure.
These findings empirically confirm that human-collected UMI data cannot be treated as directly executable robot supervision; without physical validation, VLA models would inevitably learn from physically infeasible and potentially hazardous trajectories.

\subsection{Data-Level Validation Experiments}
\label{sec:validation_effectiveness}

Having established wrist-fisheye perception mismatch and physical infeasibility as two key bottlenecks in UMI-to-VLA adaptation, we next evaluate whether the two data-level components of VISTA effectively address these issues: UMI-VQA for perception-aligned supervision and physical validation for executability-aware trajectory selection.

\paragraph{Effect of perception-aligned VQA.}
Given the wrist-fisheye perception mismatch identified in Section~\ref{sec:exp-diagnostic}, we evaluate whether UMI-VQA provides effective auxiliary supervision for policy learning under the UMI observation regime. We conduct this study on top of $\pi_{0.5}$ using a controlled subset of UMI training data. Across all variants, we keep the UMI action data, observation inputs, training budget, and optimization procedure fixed, and vary only the auxiliary VQA supervision. We compare three settings: action-only training without VQA supervision, co-training with standard-view VQA, and co-training with UMI-VQA. Here, standard-view VQA denotes auxiliary VQA supervision composed of standard-perspective web VQA data and main-view robot VQA data, whose visual distributions differ from the wrist-fisheye observations used by UMI policies. As shown in Table~\ref{tab:vqa_source_ablation}, standard-view VQA achieves a lower aggregate success rate than action-only training. This result suggests that auxiliary VQA supervision is not universally beneficial in co-training. One possible explanation is that standard-view VQA provides supervision under global, regular-perspective observations, whereas UMI action prediction relies on local, distorted, gripper-centric wrist-fisheye cues. When optimized with a shared backbone, such distributional mismatch may bias representation learning away from the visual cues needed for UMI-style action prediction. By contrast, co-training with UMI-VQA achieves the highest overall success rate across the three evaluated tasks. Compared with action-only training, UMI-VQA improves the aggregate success rate from 45.0\% to 55.0\%. The improvement is most pronounced on Stack Pen Holders, while the results on Stack Cubes and Stack Cups are comparable to action-only training. Compared with standard-view VQA, UMI-VQA also yields a higher aggregate success rate, improving from 31.7\% to 55.0\%.
The results suggest that wrist-fisheye-aligned VQA provides a more suitable auxiliary signal for UMI-style policy learning.

\begin{table}[t]
\centering
\small
\setlength{\tabcolsep}{3.5pt}
\begin{tabular}{lcccc}
\toprule
\textbf{Training Setting} 
& \textbf{Task 1} 
& \textbf{Task 2} 
& \textbf{Task 3} 
& \textbf{Overall} \\
\midrule
Action-only $\pi_{0.5}$         
& 9/20 \; (45.0\%) 
& 10/20 \; (50.0\%) 
& 8/20 \; (40.0\%) 
& 27/60 \; (45.0\%) \\

$\pi_{0.5}$ + Standard-view VQA 
& 4/20 \; (20.0\%) 
& 4/20 \; (20.0\%) 
& 11/20 \; (55.0\%) 
& 19/60 \; (31.7\%) \\

$\pi_{0.5}$ + UMI-VQA           
& 8/20 \; (40.0\%) 
& 11/20 \; (55.0\%) 
& 14/20 \; (70.0\%) 
& 33/60 \; (55.0\%) \\
\bottomrule
\end{tabular}
\caption{
Real-robot success rates under different auxiliary VQA supervision sources.
Task 1, Task 2, and Task 3 correspond to Stack Side Cubes on Center Cube, Stack Paper Cups, and Stack Pen Holders, respectively.
Each setting is evaluated over 20 trials per task, and the overall score aggregates all 60 trials.
}
\label{tab:vqa_source_ablation}
\end{table}

\paragraph{Effects of physical score validation.}
We evaluate whether physical validation scores can guide UMI trajectory filtering before policy training. As motivated by the physical gap diagnosed in Sec.~\ref{sec:exp-diagnostic}, human-collected UMI trajectories are not guaranteed to be reproducible by a target robot arm. Here, we test this risk through real-robot deployment after policy learning. We use the stapler-placement task as a controlled case study and compare policies trained on UMI subsets with different RealMan-conditioned validation scores under the same model, data budget, and training procedure. Since the main embodiment-induced execution errors in this task occur during post-grasp placement, we report Grasping Success Rate (GSR), Overall Success Rate (OSR), and Post-grasp Success Rate (PSR), computed as OSR/GSR when GSR is nonzero.

\begin{figure}[t]
    \centering
    \begin{minipage}{0.31\linewidth}
        \centering
        \includegraphics[width=\linewidth]{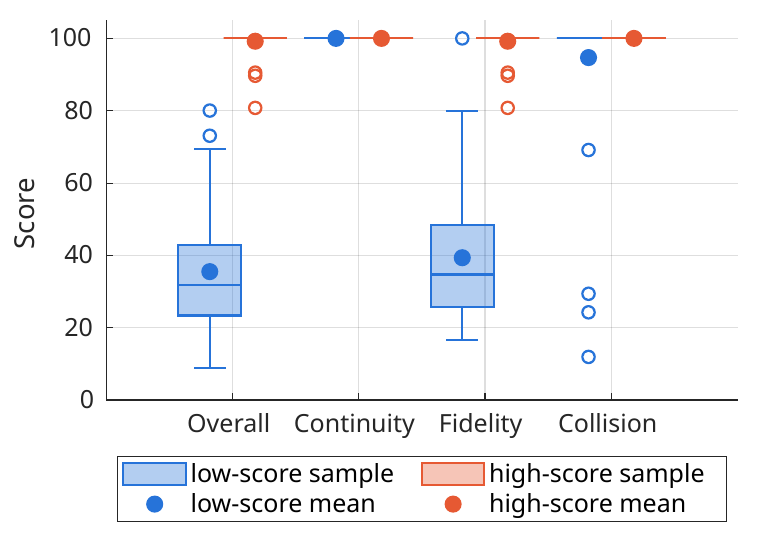}
        \vspace{0.05cm}
        {\centering (a) RealMan\par}
    \end{minipage}
    \hfill
    \begin{minipage}{0.31\linewidth}
        \centering
        \includegraphics[width=\linewidth]{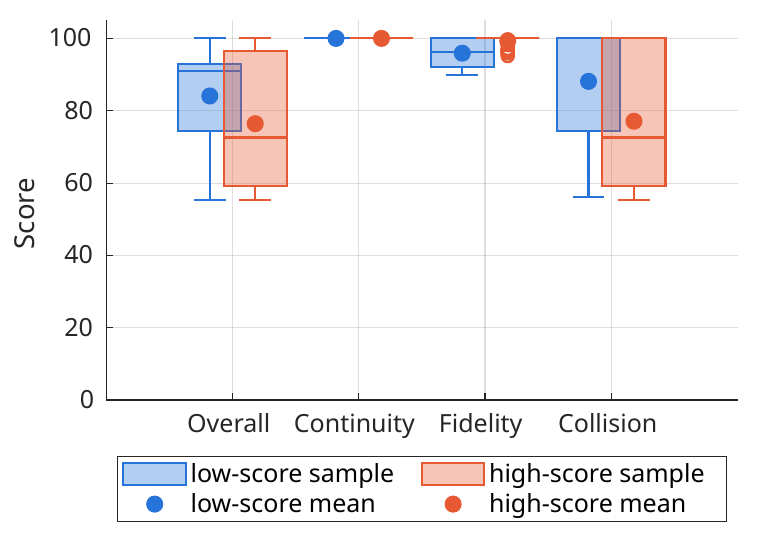}
        \vspace{0.05cm}
        {\centering (b) R1Pro\par}
    \end{minipage}
    \hfill
    \begin{minipage}{0.31\linewidth}
        \centering
        \includegraphics[width=\linewidth]{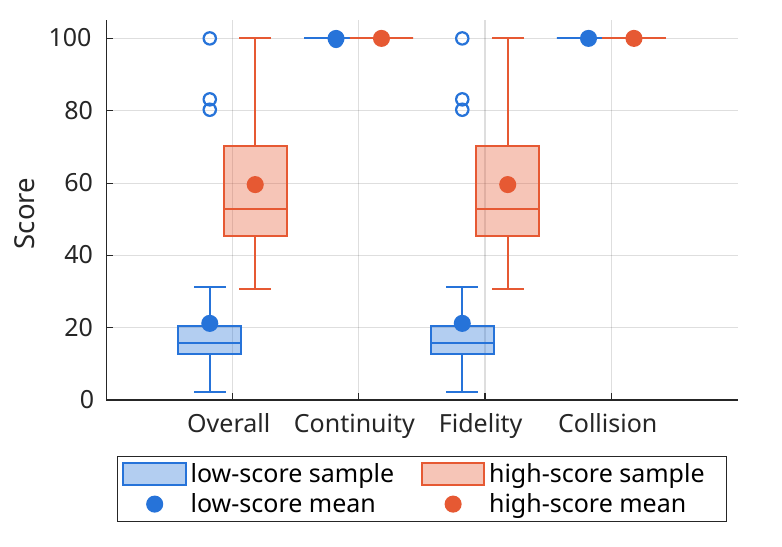}
        \vspace{0.05cm}
        {\centering (c) ACone\par}
    \end{minipage}
    \caption{
    Physical-validation score distributions of the same low-score and high-score UMI trajectory subsets across different robot embodiments. The subsets are selected on RealMan and then re-scored on R1Pro and ACone. Panel (a) shows the score separation used in the score-controlled comparison, while panels (b)--(c) show that the same subsets receive different score distributions on other embodiments, indicating that trajectory executability depends on the target robot embodiment.
    }
    \label{fig:cross_embodiment_scores}
\end{figure}

\textbf{Score-controlled comparison.}
To control for data quantity while varying target-embodiment compatibility, we construct two RealMan-conditioned subsets with the same number of demonstrations but different validation scores: a low-score subset and a high-score subset. As shown in Fig.~\ref{fig:cross_embodiment_scores}(a), the two subsets are clearly separated in their RealMan validation scores. Table~\ref{tab:score_controlled_validation} reports the corresponding deployment results. The two policies achieve comparable GSR, but differ sharply in OSR and PSR: the low-score policy can grasp the object in some trials but fails to complete post-grasp placement, whereas the high-score policy achieves much higher post-grasp and overall success. These results show that higher target-embodiment compatibility leads to more reliable real-robot deployment after policy training.

\begin{table}[t]
\centering
\small
\setlength{\tabcolsep}{4pt}
\begin{tabular}{lcccccccc}
\toprule
\textbf{Traj. Type}
& \textbf{\#Traj.}
& \textbf{Continuity}
& \textbf{Collision}
& \textbf{Fidelity}
& \textbf{Avg. Score}
& \textbf{GSR}
& \textbf{OSR}
& \textbf{PSR} \\
\midrule
Low-score subset  & 50 & 100.00 & 94.69  & 39.35 & 35.50 & 0.55 & 0.00 & 0.00 \\
High-score subset & 50 & 100.00 & 100.00 & 99.21 & 99.21 & 0.65 & 0.65 & 1.00 \\
\bottomrule
\end{tabular}
\caption{
Score-controlled subset analysis on the RealMan embodiment. Both subsets contain 50 demonstrations and are evaluated over 20 trials on the stapler-placement task with RealMan. PSR is computed as OSR/GSR and measures task completion conditioned on successful grasping.
}
\label{tab:score_controlled_validation}
\end{table}

\begin{figure}[t]
    \centering
    \includegraphics[height=3.6cm]{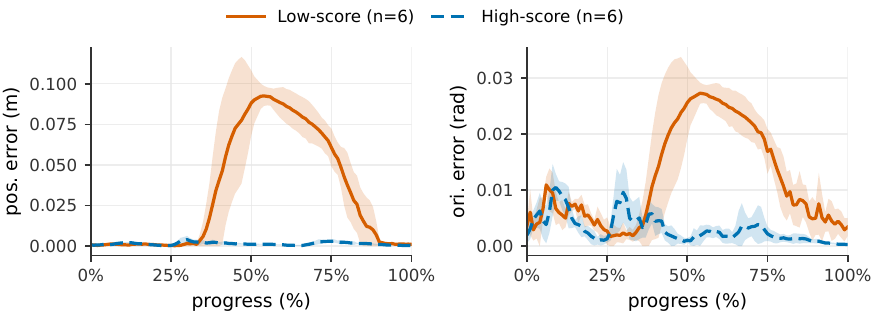}
    \caption{
Failure analysis of policies trained on low-score and high-score subsets during RealMan deployment. The curves show the mean position and orientation errors between the policy-generated desired trajectory and the nearest feasible trajectory under the target embodiment constraints, plotted over task progress. Shaded regions indicate variation across six deployment trials. The low-score policy exhibits large post-grasp deviations, indicating poor trajectory followability during placement and leading to task failure. Detailed pose trajectories are provided in Appendix~\ref{appendix:replay-details}.
    }
    \label{fig:deploy_errors}
\end{figure}

\begin{figure}[t]
    \centering
    \includegraphics[width=1\textwidth,trim=0bp 105bp 0bp 0bp]{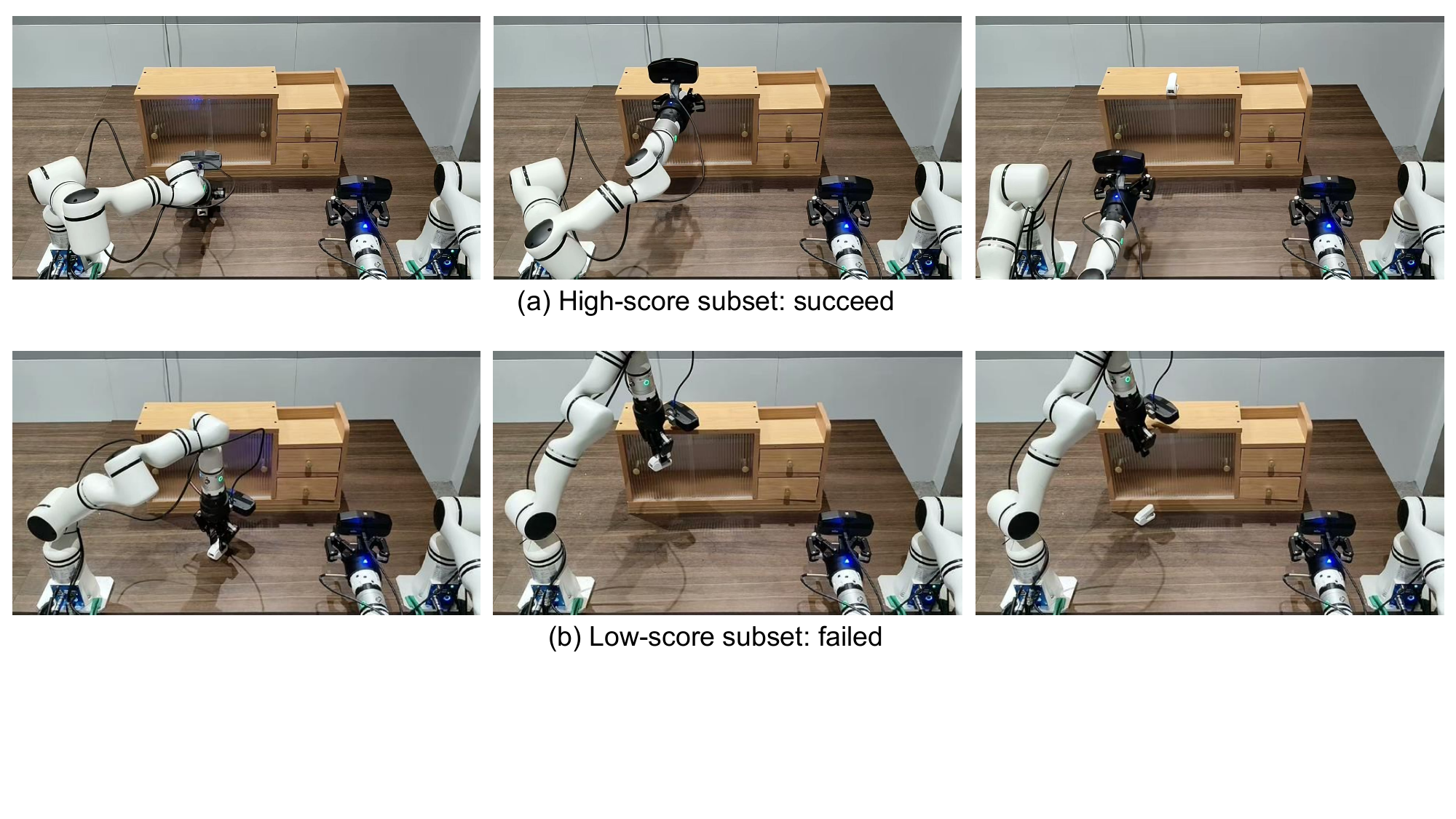}
    \caption{
    Snapshots of deployment experiments on RealMan. The low-score subset leads to post-grasp execution failure, while the high-score subset leads to successful task completion.
    }
    \label{fig:validation_snapshots}
\end{figure}

\textbf{Failure analysis.}
We further inspect representative deployment cases to understand how low-score training trajectories affect policy execution. As shown in Fig.~\ref{fig:deploy_errors}, the low-score policy generates post-grasp desired poses that are difficult for RealMan to realize, producing a large gap between the desired and nearest feasible trajectories during placement. In contrast, the high-score policy generates trajectories that can be closely followed by the target embodiment. The deployment snapshots in Fig.~\ref{fig:validation_snapshots} illustrate the resulting failure-success contrast. These results suggest that low-score training data can lead the learned policy to produce motions that are semantically plausible but physically hard for the target robot to execute.

\textbf{Embodiment-conditioned scoring.}
We next examine whether trajectory filtering should be conditioned on the target embodiment. Since the validation score is defined as \(S(\xi,e)\), the same trajectory can receive different scores under different robot embodiments. We therefore re-score the RealMan-selected low-score and high-score subsets on R1Pro and ACone. As shown in Fig.~\ref{fig:cross_embodiment_scores}, the same subsets exhibit different score distributions across embodiments. In particular, the low-score subset on RealMan receives higher scores on R1Pro, suggesting that data poorly matched to one robot may be more executable on another. The deployment results in Table~\ref{tab:multi_embodiment_validation} show the same trend: the low-score policy fails on RealMan and ACone, but achieves non-zero OSR and high PSR on R1Pro. These results support embodiment-conditioned filtering: the same UMI data can be risky for one robot but suitable for another, depending on embodiment-specific reachability and trajectory-following constraints.

\begin{table}[t]
\centering
\small
\setlength{\tabcolsep}{3.5pt}
\begin{tabular}{lcccccccccc}
\toprule
\multirow{2}{*}{\textbf{Training Data}} 
& \multirow{2}{*}{\textbf{\#Demos}} 
& \multicolumn{3}{c}{\textbf{RealMan}} 
& \multicolumn{3}{c}{\textbf{ACone}} 
& \multicolumn{3}{c}{\textbf{R1Pro}} \\
\cmidrule(lr){3-5} \cmidrule(lr){6-8} \cmidrule(lr){9-11}
& & \textbf{GSR} & \textbf{OSR} & \textbf{PSR}
  & \textbf{GSR} & \textbf{OSR} & \textbf{PSR}
  & \textbf{GSR} & \textbf{OSR} & \textbf{PSR} \\
\midrule
Low-score subset  & 50 & 0.55 & 0.00 & 0.00 & 0.60 & 0.00 & 0.00 & 0.80 & 0.80 & 1.00 \\
High-score subset & 50 & 0.65 & 0.65 & 1.00 & 0.60 & 0.55 & 0.92 & 0.75 & 0.75 & 1.00 \\
\bottomrule
\end{tabular}
\caption{
Cross-embodiment deployment results for policies trained with low-score and high-score UMI data subsets. Each policy is evaluated over 20 trials on the stapler-placement task under the same deployment protocol.
}
\label{tab:multi_embodiment_validation}
\end{table}

Together, these real-robot results validate physical scoring as a practical criterion for UMI data curation. Training directly on unvalidated UMI trajectories can introduce embodiment-mismatched supervision, while target-embodiment-conditioned filtering selects trajectories that the deployment robot is more likely to reproduce reliably.

\subsection{Model Evaluation Experiments}
\label{sec:policy_results}

After validating the roles of perception-aligned auxiliary supervision and executability-aware trajectory selection, we further evaluate the complete VISTA pipeline under UMI-style settings.

\paragraph{Simulation benchmark.}
We evaluate VISTA against three strong VLA baselines on our adapted UMI-style simulation benchmark, including RoboTwin-UMI and LIBERO-UMI.
All methods are trained on the same recollected wrist-fisheye demonstrations, ensuring that the comparison isolates the effect of the proposed UMI-oriented adaptation rather than differences in training data.
As shown in Table~\ref{tab:sim_main_results}, VISTA achieves the best performance on both benchmarks.
On RoboTwin-UMI, VISTA improves the success rate from 0.594 to 0.683 over the strongest baseline $\pi_{0.5}$.
On LIBERO-UMI, VISTA further improves from 0.922 to 0.943.
Overall, VISTA obtains an average success rate of 0.813, outperforming $\pi_{0.5}$ by 5.5 points, LingBot-VLA by 15.5 points, and Wall-X by 38.7 points.
These results show that explicitly adapting the model to UMI-style wrist-fisheye observations and physically validated action data leads to more effective policy learning under controlled same-data conditions.

\begin{table}[t]
\centering
\small
\begin{tabular}{lccc}
\toprule
\textbf{Model} & \textbf{RoboTwin-UMI} & \textbf{LIBERO-UMI} & \textbf{Avg.} \\
\midrule
LingBot-VLA       & 0.499    & 0.817 & 0.658    \\
Wall-X        & 0.152 & 0.700 & 0.426 \\
$\pi_{0.5}$   & 0.594 & 0.922 & 0.758 \\
VISTA     & 0.683 & 0.943 & 0.813 \\
\bottomrule
\end{tabular}
\caption{
Main simulation results under UMI-style wrist-fisheye observations.
All methods are trained on the same recollected demonstrations. 
}
\label{tab:sim_main_results}
\end{table}

\begin{wraptable}{r}{0.42\textwidth}
\vspace{-1.0em}
\centering
\small
\begin{tabular}{lc}
\toprule
\textbf{Policy} & \textbf{Avg. Success Rate} \\
\midrule
LingBot-VLA & 0.313 \\
$\pi_{0.5}$ & 0.528 \\
VISTA & \textbf{0.598} \\
\bottomrule
\end{tabular}
\caption{
Real-robot evaluation averaged over 20 UMI-collected manipulation tasks.
All methods are trained on the same validated UMI dataset and evaluated with 20 trials per task.
Success rates are reported in $[0,1]$.
}
\label{tab:real_results_avg}
\vspace{-1.0em}
\end{wraptable}

\paragraph{Real-robot evaluation.}
We further evaluate VISTA on 20 real-world UMI-collected manipulation tasks, covering precise spatial localization, dual wrist-view integration, and local interaction reasoning.
All methods are trained on the same validated UMI dataset and evaluated under the same robot platform, task setup, and object configurations.
For each task, we conduct 20 real-world trials and report the average success rate across all tasks.
As shown in Table~\ref{tab:real_results_avg}, VISTA achieves the highest average success rate among all compared methods.
It improves over $\pi_{0.5}$ from 0.528 to 0.598, corresponding to a 7.0-point absolute gain, and substantially outperforms LingBot-VLA by 28.5 points.
These results demonstrate that the benefits of VISTA transfer beyond simulation to real-world deployment, where wrist-fisheye perception, local interaction understanding, and physically executable action supervision are all critical.
Detailed task descriptions and per-task results are provided in Appendix~\ref{app:more_exp_details}.


\subsection{Ablations and Analysis}
\label{sec:ablation_analysis}

We finally analyze which components of VISTA contribute to the final performance.
We study model-level components and action/state representation choices.

\paragraph{Component ablation.}
We conduct a unified component ablation to analyze the key design choices of VISTA.
As shown in Table~\ref{tab:component_ablation}, the full model achieves a success rate of 68.3\%.
Removing the Stage-2 training design substantially degrades performance: using a scratch expert drops success to 52.4\%, while replacing it with the original $\pi_{0.5}$ expert reaches only 60.2\%.
This confirms that the Stage-2 expert learned in VISTA provides a stronger action prior for UMI-style wrist-fisheye policies. We also find that both state conditioning and delta action prediction are important.
Removing proprioceptive state input reduces success to 61.9\%, indicating that embodiment and execution context complements visual observations.
Replacing delta action prediction with absolute action prediction further reduces success to 53.1\%, suggesting that relative action representation is crucial for mapping handheld UMI trajectories to robot execution.
Together, these ablations validate the contributions of the learned action expert, proprioceptive state input, and delta action representation.

\begin{table}[t]
\centering
\small
\setlength{\tabcolsep}{4pt}
\begin{tabular}{lcccc}
\toprule
\textbf{Variant} 
& \textbf{Stage-2 Expert} 
& \textbf{State} 
& \textbf{Delta Action} 
& \textbf{Success} \\
\midrule
VISTA 
& ours 
& \checkmark 
& \checkmark 
& 68.3 \\

w/o Stage 2, scratch expert
& scratch 
& \checkmark 
& \checkmark 
& 52.4 \\

w/o Stage 2, $\pi_{0.5}$ expert
& $\pi_{0.5}$ 
& \checkmark 
& \checkmark 
& 60.2 \\

w/o state
& ours 
& -- 
& \checkmark 
& 61.9 \\

w/o delta action
& ours 
& \checkmark 
& -- 
& 53.1 \\
\bottomrule
\end{tabular}
\caption{
Unified component ablation of VISTA.
We compare the full model with variants that replace the Stage-2 action expert, remove proprioceptive state input, or remove delta action prediction.
The ablation tests the contributions of the learned action expert, state conditioning, and relative action representation.
}
\label{tab:component_ablation}
\end{table}

\paragraph{Attention-based analysis of visual grounding.}

\begin{figure*}[t]
    \centering
    \includegraphics[width=0.75\textwidth]{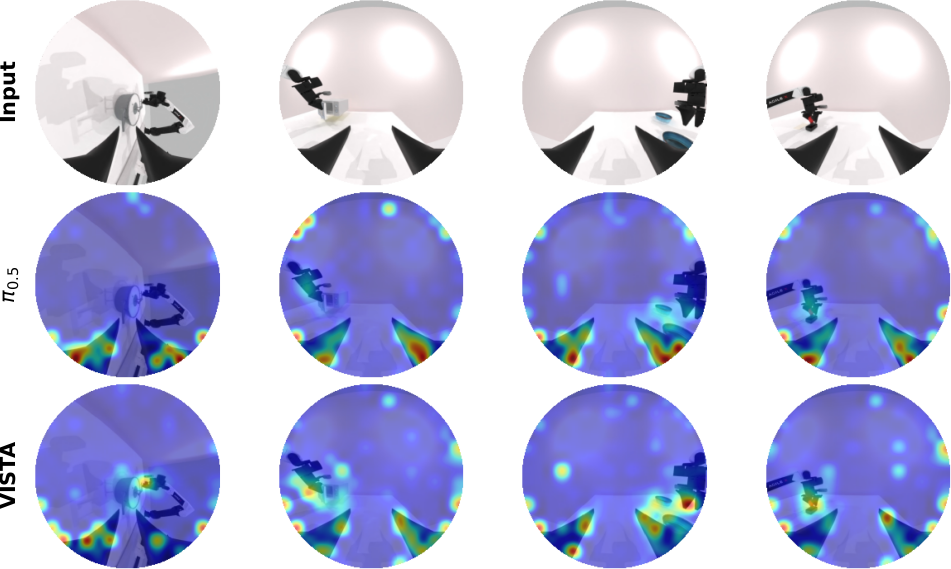}
    \caption{
    Attention visualization under dual UMI-style wrist-fisheye observations.
    Columns correspond to lift pot, open microwave, stack bowls, and stamp seal, respectively.
    Rows show the input image, $\pi_{0.5}$ attention, and VISTA attention.
    }
    \label{fig:attention_visualization}
\end{figure*}

We further visualize attention maps to qualitatively inspect how VISTA processes dual wrist-fisheye observations. The visualization compares $\pi_{0.5}$ with VISTA on representative RoboTwin-UMI frames. Under this observation regime, the policy relies on wrist-mounted fisheye cameras rather than a global main-view camera, and the two wrist views can provide complementary information about the active gripper, target object, and local interaction region. As shown in Fig.~\ref{fig:attention_visualization}, each example is taken from the wrist camera of the non-acting gripper, which serves as an auxiliary observer of the manipulation performed by the other gripper. Compared with $\pi_{0.5}$, which often exhibits more diffuse attention in this auxiliary wrist view, VISTA tends to produce more localized attention around task-relevant regions, including the active gripper, manipulated object, and local interaction area. This qualitative comparison suggests that VISTA develops stronger visual grounding under UMI-style observations.

\section{Conclusion}

In this work, we presented VISTA, a vision-grounded and physics-validated framework for adapting UMI-collected data to VLA training. We identified two key mismatches in raw UMI data: wrist-mounted fisheye observations are visually misaligned with standard VLM pretraining domains, and human-collected trajectories may be physically infeasible for downstream robot embodiments. VISTA addresses these issues through UMI-VQA for fisheye-aligned visual grounding, trajectory-level physical validation for embodiment-compatible data curation, and a two-stage co-training recipe for vision-language-action learning. Experiments on UMI-style simulation benchmarks, real-world manipulation tasks, and cross-embodiment deployment show that VISTA consistently improves over strong VLA baselines. These results suggest that explicit perceptual alignment and physical feasibility validation are essential for scaling VLA training with handheld demonstration data.

\clearpage

\bibliographystyle{plainnat}
\bibliography{paper}

\clearpage
\appendix
\section{The details of Multi-robot deployment System}
\label{appendix:multi-deployment}

\begin{figure}[h!]
    \centering
    \includegraphics[width=1.0\textwidth]{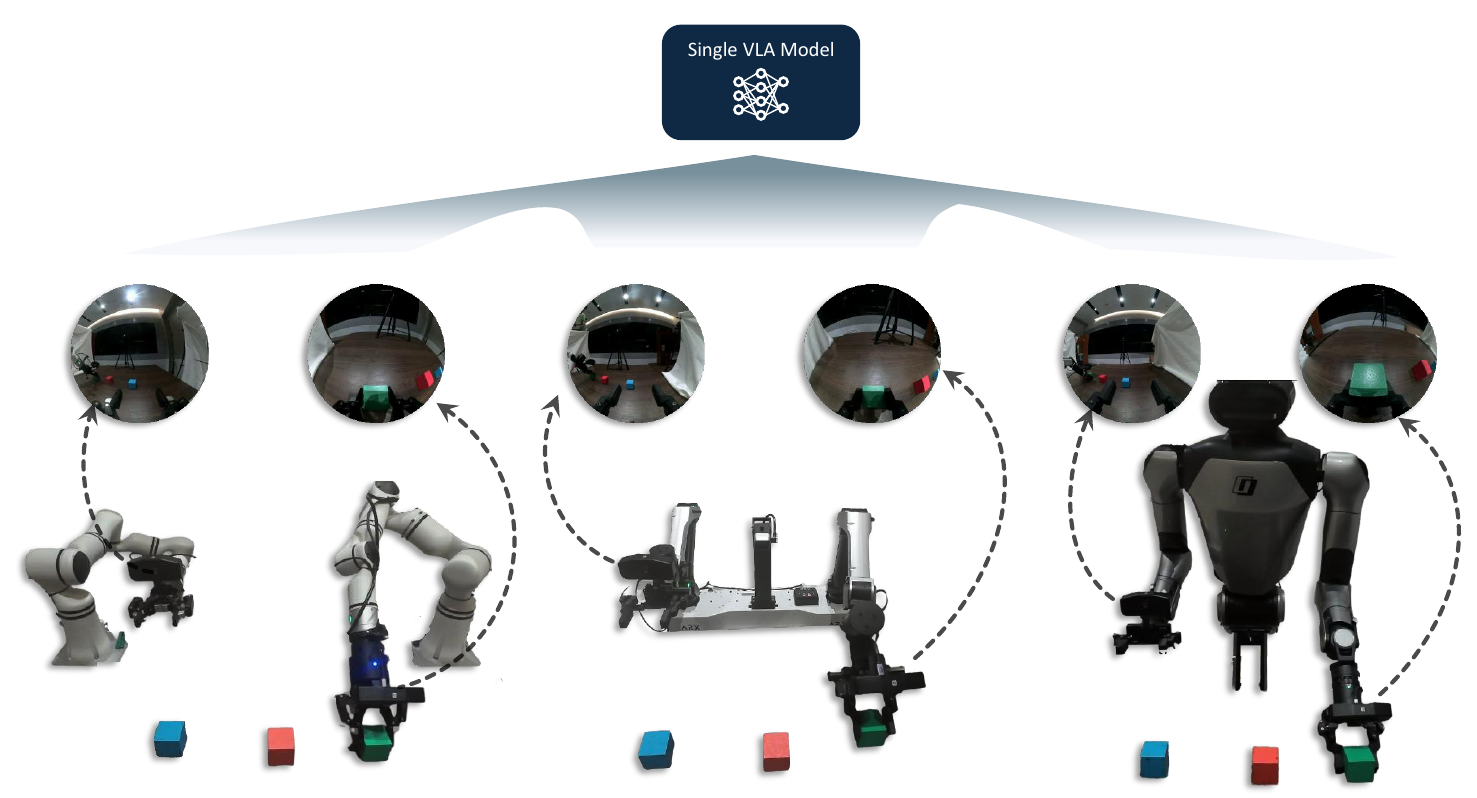}
    \caption{As shown in the figure, this is a schematic diagram of us using the same model to control heterogeneous robotic arms to complete the same task. By using our framework, a single model can be deployed simultaneously on these heterogeneous robotic arms.}
    \label{fig:multireal}
\end{figure}

The primary engineering objective of this work is to fully exploit the deployment potential of UMI-pretrained VLA models on heterogeneous devices. To this end, we design and implement a pure Python-based distributed deployment architecture for multiple heterogeneous robotic arms. This architecture eliminates the heavy reliance on the ROS ecosystem typical in traditional embodied AI deployments, thereby reducing complex learning and configuration overheads. Meanwhile, by employing Zenoh as the underlying communication middleware, the system automatically adapts to the communication medium (inter-process on the same machine, LAN, or even WAN) to achieve shared-memory-level zero-copy or transparent network-level transmission. This design circumvents the concurrency bottlenecks caused by the GIL in older Python versions (<3.13), maximizing the utilization of all available computational resources within the system.

Benefiting from the inherent uniformity of the UMI dataset in both state observation and action spaces, our system has achieved a high degree of compatibility with heterogeneous robotic arms. In practical deployment, for any unseen heterogeneous robotic arm, provided that the physical mounting of its end-effector and camera aligns with UMI data collection specifications, developers should only need to implement a basic Cartesian space control interface based on "XYZ spatial coordinates + quaternions." Once this step is completed, the heterogeneous robotic arm can be seamlessly integrated into the current scheduling network as a standard physical terminal, requiring no additional custom adaptations on the algorithm side or other architectural components.

As illustrated in Fig.\ref{fig:multi2multi}, implementation follows a strictly decoupled host-satellite logic. The distributed compute nodes—depicted as Satellite Machine 1 through N in the lower tier—run persistently as lightweight system-level daemons. During bootstrapping, the Host Machine simply triggers its master orchestrator, which dispatches a configuration JSON payload via the Zenoh backbone to the target satellites. Upon receipt, these satellite orchestrators autonomously instantiate their respective VLA Model and data processing pipelines without any manual intervention. This elegantly reduces the deployment of a complex multi-machine, multi-GPU cluster across a LAN to the execution of a single local script, achieving absolute transparency and zero redundant operational overhead.

At the data flow and inference layer, independent state streams from each robotic arm are aggregated on the computing nodes via the Zenoh network. Notably, when evaluating different embodied policy models, researchers only need to modify the corresponding model inference process configuration in a YAML file, while the entire distributed communication architecture remains completely unchanged. Furthermore, the multi-machine asynchronous inference mechanism eliminates the computational and I/O blocking typical of single nodes, ensuring high-frequency control responses.

During the action dispatch phase, the batched predicted actions output by all parallel computing nodes ultimately converge at the core "Ensembling and Routing Node." In our implementation, we utilize Temporal Ensembling to leverage and fuse the action chunks generated by the multi-machine asynchronous inference. Once the action trajectories undergo temporal fusion, the Router performs targeted distribution based on a predefined mapping table. In this critical process, the RPC (Remote Procedure Call) mechanism plays an essential, interlocking role: rather than dispatching commands via a loosely coupled publish/subscribe pattern, the Temporal Ensembling node employs RPC to synchronously call the underlying robot control processes. It strictly ensures that the corresponding commands have been pushed into the command queue of the robot control process before proceeding to unpack and dispatch the next frame of actions. This strongly coupled state machine design based on RPC ensures strict temporal synchronization between the compute cluster and the physical execution terminals, thereby preventing command accumulation or frame dropping. As illustrated in Fig. \ref{fig:multireal}, this robust framework ultimately enables a single model to be deployed simultaneously across three distinct machines to control heterogeneous robotic arms for the same task.

Taken together, this engineering implementation fully utilizes the inherent state-alignment advantages of VISTA and instantiates the AI Flow vision in heterogeneous embodied AI systems, demonstrating a deployment paradigm from distributed inference clusters to heterogeneous robot clusters\citep{aiflow}

\section{Details of poses in the replay and deployment}
\label{appendix:replay-details}

\begin{figure}[t]
    \centering
    \begin{minipage}{0.48\linewidth}
        \centering
        \includegraphics[width=\linewidth]{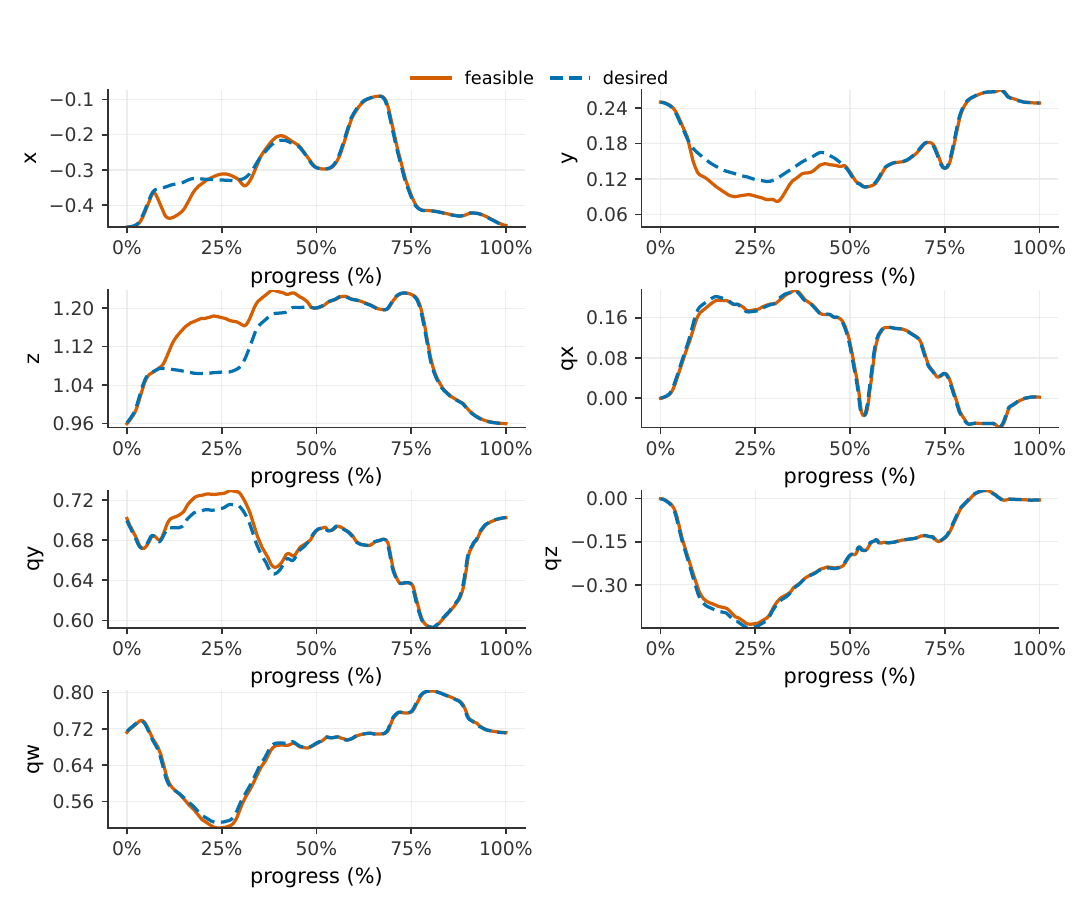}
        \vspace{0.1cm}
        {(1)Infeasible trajectory with high deviation.}
    \end{minipage}
    \hfill
    \begin{minipage}{0.48\linewidth}
        \centering
        \includegraphics[width=\linewidth]{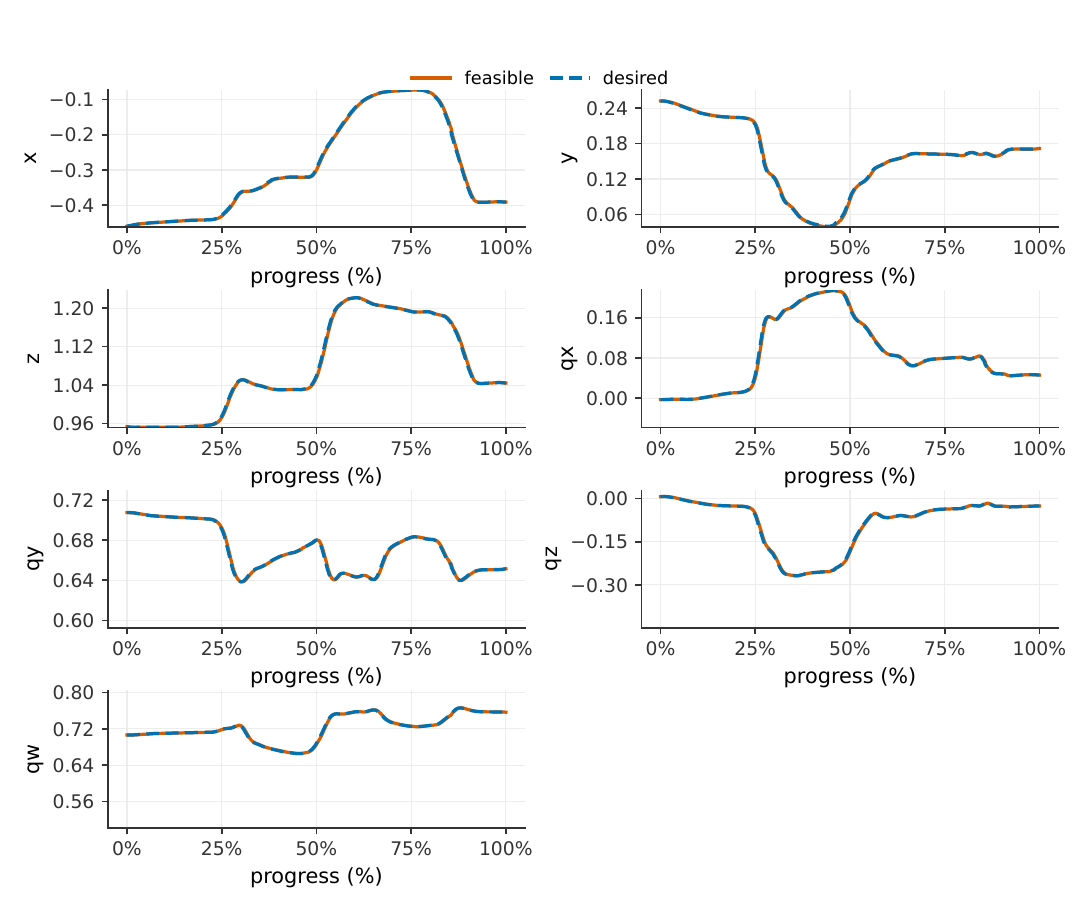}
        \vspace{0.1cm}
        {(2)Feasible trajectory with low deviation.}
    \end{minipage}
    \caption{
    Desired and feasible poses observed during the replay of UMI trajectories on the RealMan robot executing the glue stick handover task.
    }
    \label{fig:ee_tracking hand_over_glue_stick}
\end{figure}

\begin{figure}[t]
    \centering
    \begin{minipage}{0.48\linewidth}
        \centering
        \includegraphics[width=\linewidth]{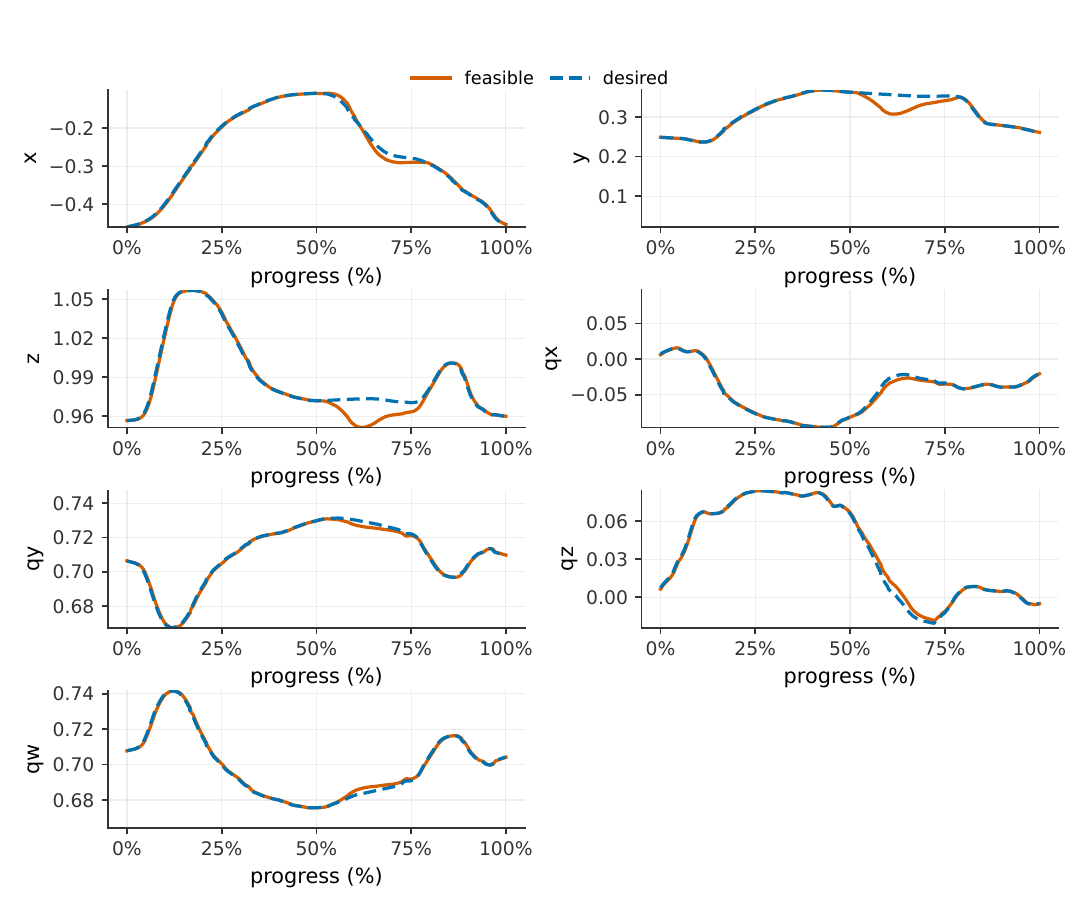}
        \vspace{0.1cm}
        {(1)Infeasible trajectory with high deviation.}
    \end{minipage}
    \hfill
    \begin{minipage}{0.48\linewidth}
        \centering
        \includegraphics[width=\linewidth]{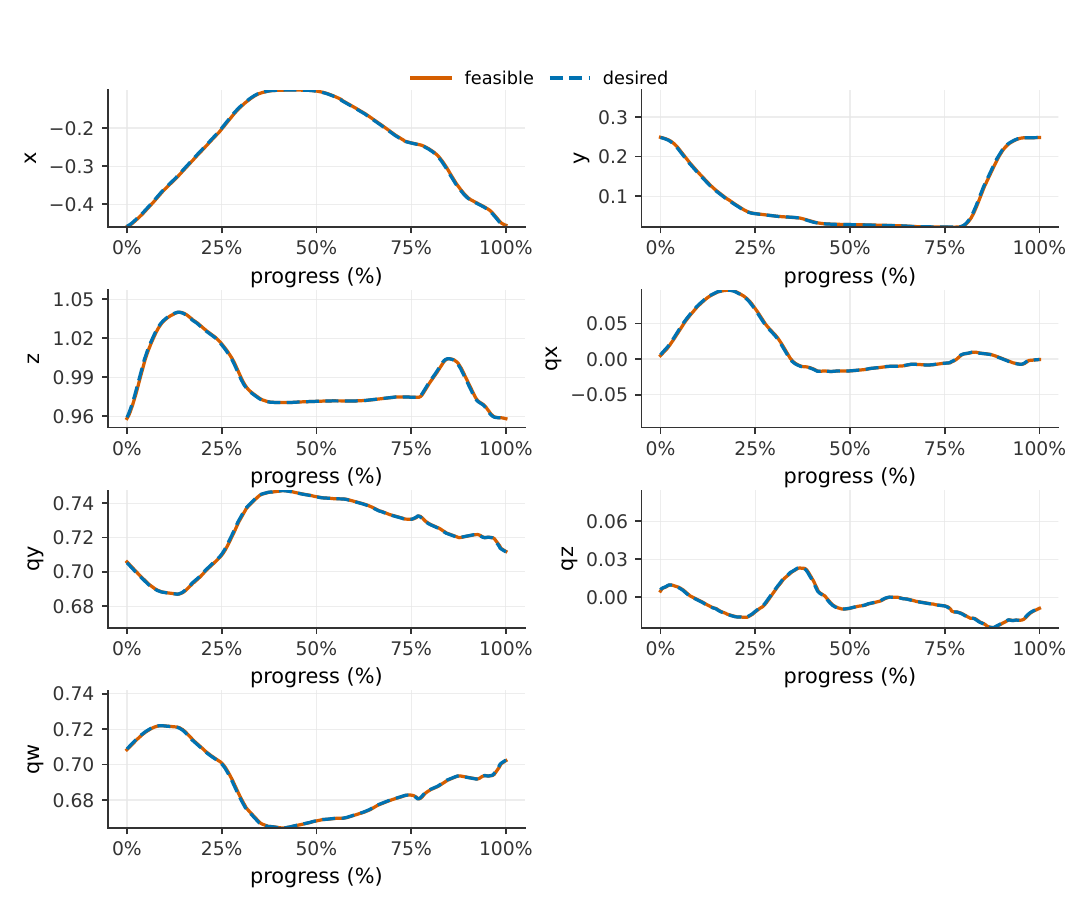}
        \vspace{0.1cm}
        {(2)Feasible trajectory with low deviation.}
    \end{minipage}
    \caption{
    Desired and feasible poses observed during the replay of UMI trajectories on the RealMan robot executing the drawer-pulling task.
    }
    \label{fig:ee_tracking pull_drawer}
\end{figure}

\begin{figure}[t]
    \centering
    \begin{minipage}{0.48\linewidth}
        \centering
        \includegraphics[width=\linewidth]{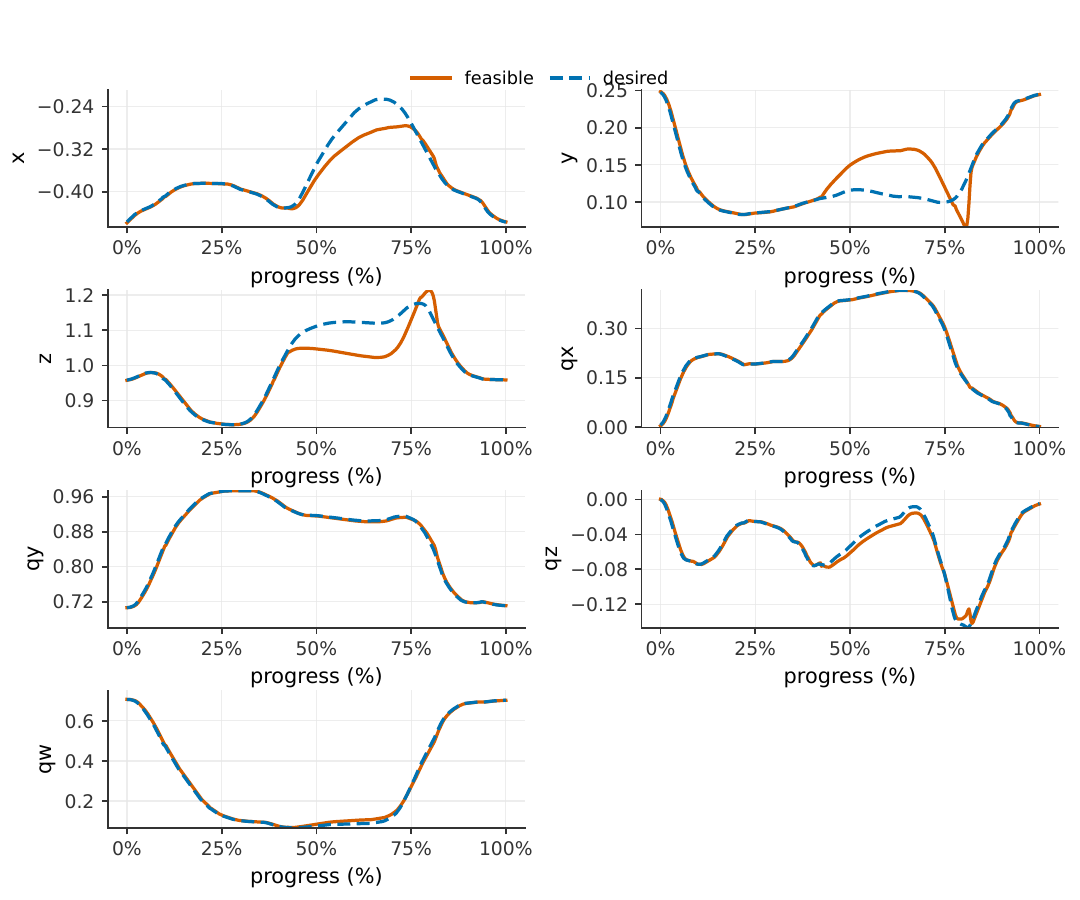}
        \vspace{0.1cm}
        {(1)Infeasible trajectory with high deviation.}
    \end{minipage}
    \hfill
    \begin{minipage}{0.48\linewidth}
        \centering
        \includegraphics[width=\linewidth]{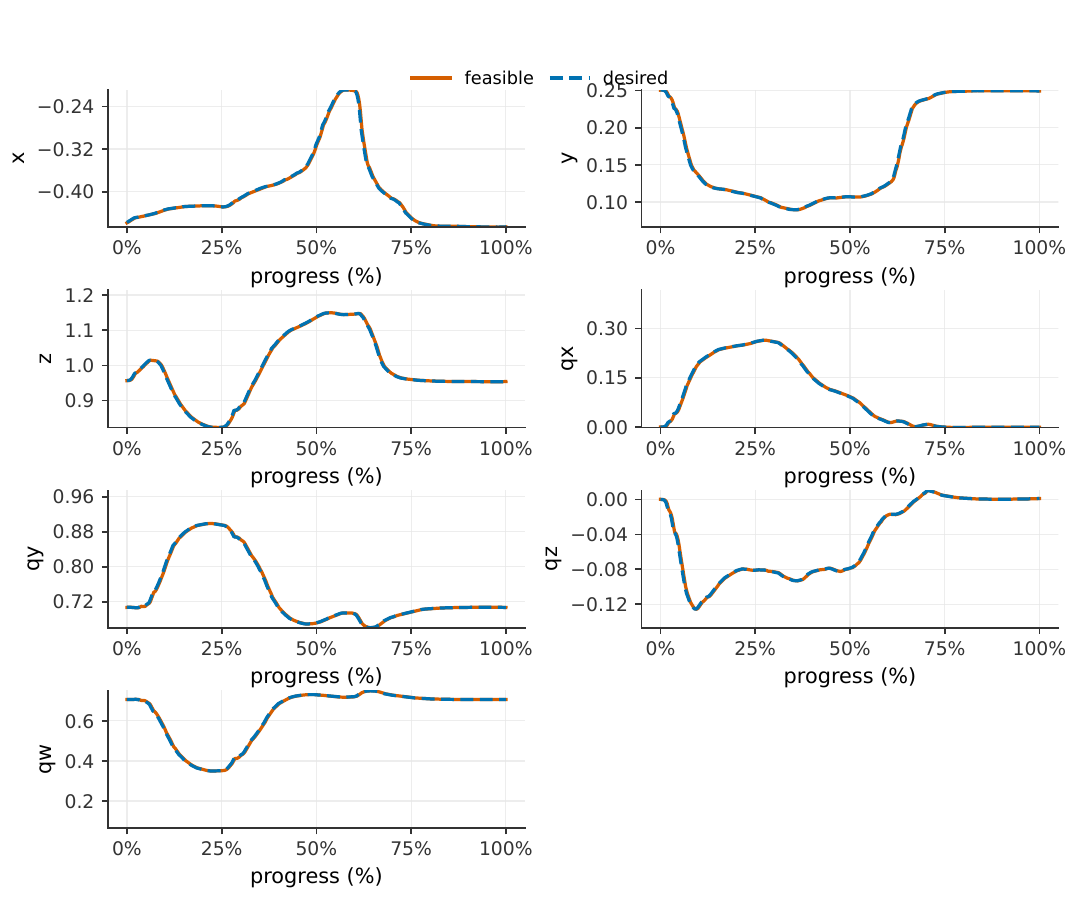}
        \vspace{0.1cm}
        {(2)Feasible trajectory with low deviation.}
    \end{minipage}
    \caption{
    Desired and feasible poses observed during the replay of UMI trajectories on the RealMan robot executing the stapler-placement task.
    }
    \label{fig:ee_tracking put_stapler_onto_shelf}
\end{figure}

\begin{figure}[t]
    \centering
    \begin{minipage}{0.48\linewidth}
        \centering
        \includegraphics[width=\linewidth]{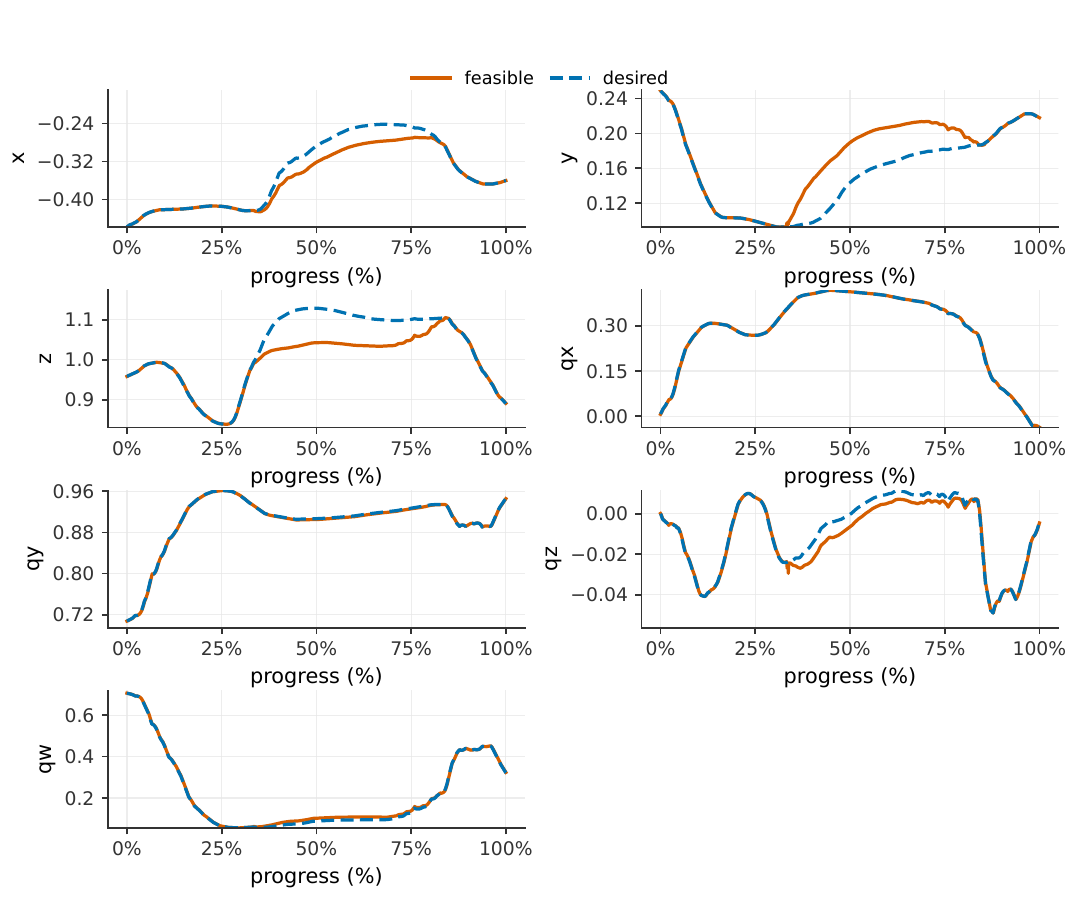}
        \vspace{0.1cm}
        {Low-score: high deviation between desired and feasible poses.}
    \end{minipage}
    \hfill
    \begin{minipage}{0.48\linewidth}
        \centering
        \includegraphics[width=\linewidth]{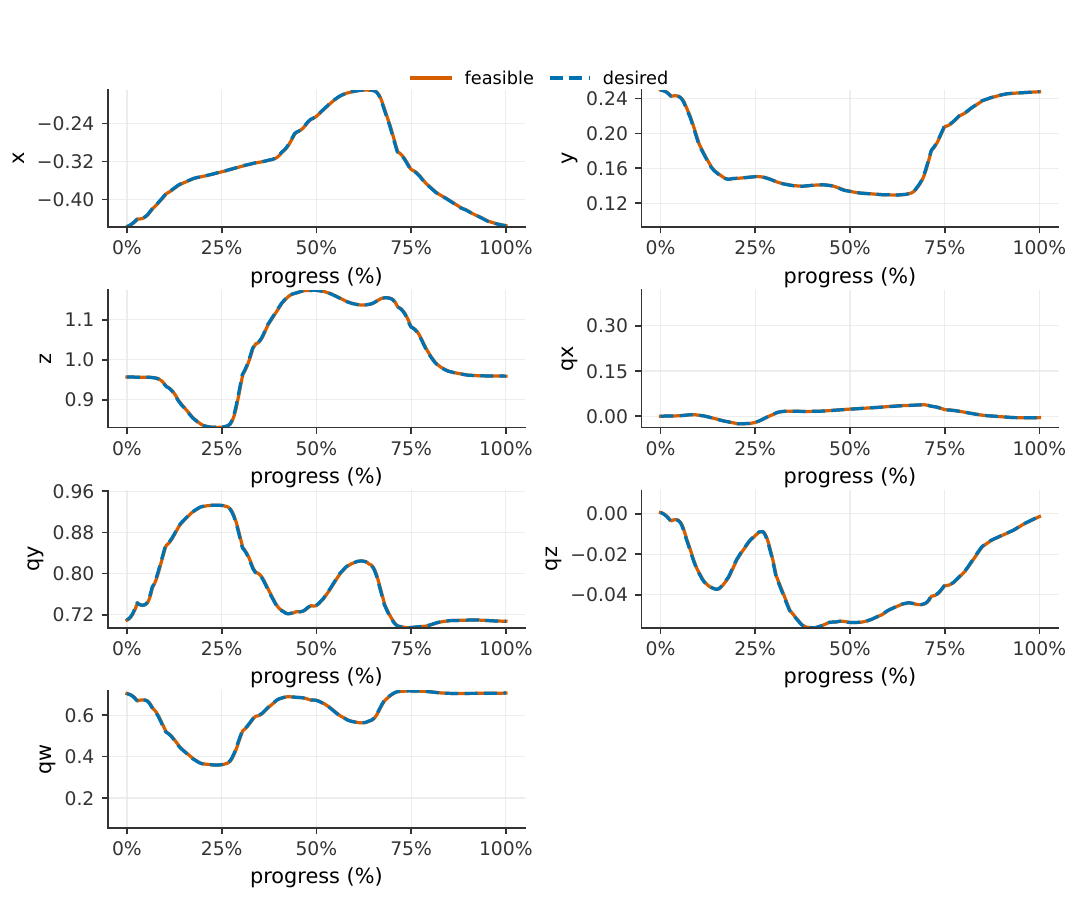}
        \vspace{0.1cm}
        {High-score: low deviation between desired and feasible poses.}
    \end{minipage}
    \caption{
    Desired and feasible end-effector poses for one deployment trail on RealMan for the stapler placement task.
    Low-score data causes tracking failure due to workspace limits, while high-score data produces more executable desired poses.
    }
    \label{fig:ee_tracking_case}
\end{figure}

Details of the desired and feasible poses observed during the replay of UMI trajectories on RealMan for the three tasks are shown in Figs.~\ref{fig:ee_tracking hand_over_glue_stick}–~\ref{fig:ee_tracking put_stapler_onto_shelf}, respectively. Orientation is represented using quaternions. Additionally, details of the desired and feasible poses for one deployment trail on RealMan for the stapler placement task are shown in Fig.\ref{fig:ee_tracking_case}. These figures more clearly demonstrate the deviation in each dimension. 

\clearpage
\section{UMI-VQA Construction Pipeline and Prompt Templates}
\label{app:umi_vqa_construction}

This appendix provides the data construction pipeline and prompt templates for UMI-VQA. UMI-VQA is constructed from two complementary sources: real-world UMI wrist-fisheye observations and a RefSpatial-based spatial-diversity supplement. The former provides authentic gripper-centric fisheye observations from manipulation demonstrations, while the latter increases the diversity of spatial relations and scene layouts under fisheye-style views.

\subsection{Real-world UMI VQA Construction}
\label{app:realworld_umi_vqa}

For the real-world UMI portion, we sample images from UMI demonstration episodes collected with paired left- and right-wrist fisheye cameras. For each episode, we randomly sample frames from both wrist views and organize the sampled images into a candidate image pool. To reduce redundancy, we remove near-duplicate frames that correspond to visually similar states within the same episode or across adjacent timesteps. Each retained wrist-fisheye image is then annotated with Qwen3-VL-235B-A22B-Instruct using five groups of prompts, corresponding to scene captioning, scene-state understanding, object grounding, interaction grounding, and spatial reasoning. For each image, we provide the image and its corresponding task instruction to the model. The task instruction is used only as a weak reference, since human-annotated task names may be noisy, incomplete, or unavailable. After generation, we post-process the resulting question--answer pairs by filtering malformed outputs and removing duplicated or near-duplicated QA pairs. This process yields the real-world UMI VQA portion organized into five capability-oriented subsets.


\begin{tcolorbox}[
    breakable,
    width=\linewidth,
    colback=orange!3,
    colframe=orange!75!black,
    coltitle=white,
    colbacktitle=orange!85!black,
    title=Prompt for Scene Captioning,
    fonttitle=\bfseries\large,
    boxrule=0.8pt,
    arc=3pt,
    left=10pt,
    right=10pt,
    top=8pt,
    bottom=8pt
]
{\ttfamily\fontsize{9pt}{10.5pt}\selectfont
\setlength{\parindent}{0pt}
\setlength{\parskip}{3pt}

You are an expert in visual analysis. Your task is to generate a concise image caption, under 80 words, for an image of a robotic arm performing a specific task.

Additional Rules:

1. Do not describe motion. Provide only a static description of the scene.\\
2. Focus only on objects that appear on the table surface or in the manipulation workspace; ignore walls, lighting, and camera viewpoint.\\
3. The terms front, back, left, right, up, and down in the generated caption should correspond to the 3D space of the image.

Input task performed by the robotic arm: \{task\_text\}

Output caption:
}
\end{tcolorbox}


\begin{tcolorbox}[
    breakable,
    width=\linewidth,
    colback=orange!3,
    colframe=orange!75!black,
    coltitle=white,
    colbacktitle=orange!85!black,
    title=Prompt for Scene-State Understanding,
    fonttitle=\bfseries\large,
    boxrule=0.8pt,
    arc=3pt,
    left=10pt,
    right=10pt,
    top=8pt,
    bottom=8pt
]
{\ttfamily\fontsize{9pt}{10.5pt}\selectfont
\setlength{\parindent}{0pt}
\setlength{\parskip}{3pt}

You are an AI assistant analyzing robot arm camera images and task instructions. Given the robot arm camera image and the task instruction: \{task\_text\}, generate natural open-ended question-answer pairs about the image and task.

Question Categories: 

1. Visible objects in the image, or the current state or position of the robot gripper.\\
2. How to accomplish the given task, or what obstacles or challenges may affect task execution.\\
3. Safety considerations or manipulation constraints relevant to the image and task.\\
4. Counterfactual reasoning: if a factor changes, such as the environment, object existence or condition, task goal, or robot condition, how must the robot adapt its movement to still complete the task?

Example: If the table is wet, can the current trajectory still complete the task?\\
Example: To grasp a durian instead of a cube, how must the robot's action change?

Answer Relevance [STRICT]:

1. Answer only the specific question asked, without adding task background, general advice, future steps, or unrelated observations.\\
2. Keep the answer scoped to the question type: if the question asks about distance, answer only about distance; if it asks about orientation, answer only about orientation.\\
3. Do not use expansion phrases such as "additionally", "furthermore", "also note", or similar expressions.\\
4. Interpret front, back, left, right, up, and down according to the 3D space of the image.

Length Constraint [STRICT]:

Question: \textless{} 50 words.\\
Answer: \textless{} 100 words.

Output Format:

Return the result as a JSON array ONLY. Do not include any explanations or extra text:

[\\
\hspace*{1em}\{\\
\hspace*{2em}"Question": "Your question here",\\
\hspace*{2em}"Answer": "Your answer here - STRICTLY LIMITED to the question scope"\\
\hspace*{1em}\},\\
\hspace*{1em}\{\\
\hspace*{2em}"Question": "Your question here",\\
\hspace*{2em}"Answer": "Your answer here - STRICTLY LIMITED to the question scope"\\
\hspace*{1em}\},\\
\hspace*{1em}...\\
]

Generate exactly 4 question-answer pairs. Make sure each question is natural and each answer is informative for understanding the robot arm's environment and task.
}
\end{tcolorbox}


\begin{tcolorbox}[
    breakable,
    width=\linewidth,
    colback=orange!3,
    colframe=orange!75!black,
    coltitle=white,
    colbacktitle=orange!85!black,
    title=Prompt for Object Grounding,
    fonttitle=\bfseries\large,
    boxrule=0.8pt,
    arc=3pt,
    left=10pt,
    right=10pt,
    top=8pt,
    bottom=8pt
]
{\ttfamily\fontsize{9pt}{10.5pt}\selectfont
\setlength{\parindent}{0pt}
\setlength{\parskip}{2pt}

You are an AI assistant specializing in visual grounding analysis of robot arm camera images.

Given the robot arm camera image and the task instruction: \{task\_text\}.

Generate diverse question-answer pairs focused on visual grounding.

Question Types:

Type 1: Direct Object Localization.\\
Target the object directly by its name.\\
Example: Detect the [object] in the image.\\
Example: Detect the [object] in the scene.

Type 2: Spatial Relationship Reasoning.\\
Target the object by describing its spatial relationship to other objects, without mentioning the target object's name.\\
Example: Detect the object positioned between [object A] and [object B].

Type 3: Task-Contextual Grounding.\\
Target the object according to the current task context \{task\_text\}, without explicitly naming the object.\\
Example: Detect the object that should be manipulated to complete the task.

Additional Rules:

1. Generate exactly 10 QA pairs: 5 questions for Type 1, 3 questions for Type 2, and 2 questions for Type 3.\\
2. All questions must start with "Detect" while maintaining natural fluency and diverse expressions.\\
3. The answers should cover both single-object and multi-object detection cases.\\
4. Interpret spatial relations, including up/down, front/back, left/right, and near/far, according to the 3D space of the image. Do not confuse front/back depth relations with up/down vertical relations.\\
5. Each answer should contain only bounding-box coordinates [x1, y1, x2, y2] and a simple object description. Do not include any reasoning process.

Output Format:

Return the result as a JSON array ONLY. Do not include any explanations or extra text:

[\\
\hspace*{1em}\{\\
\hspace*{2em}"Question": "Your visual grounding question here",\\
\hspace*{2em}"Answer": [\\
\hspace*{3em}\{"bbox\_2d": [x1, y1, x2, y2], "label": "object description"\},\\
\hspace*{3em}\{"bbox\_2d": [x1, y1, x2, y2], "label": "object description"\}\\
\hspace*{2em}]\\
\hspace*{1em}\},\\
\hspace*{1em}\{\\
\hspace*{2em}"Question": "Your visual grounding question here",\\
\hspace*{2em}"Answer": [\\
\hspace*{3em}\{"bbox\_2d": [x1, y1, x2, y2], "label": "object description"\}\\
\hspace*{2em}]\\
\hspace*{1em}\}\\
]

Generate exactly 10 question-answer pairs.
}
\end{tcolorbox}


\begin{tcolorbox}[
    breakable,
    width=\linewidth,
    colback=orange!3,
    colframe=orange!75!black,
    coltitle=white,
    colbacktitle=orange!85!black,
    title=Prompt for Interaction Grounding,
    fonttitle=\bfseries\large,
    boxrule=0.8pt,
    arc=3pt,
    left=10pt,
    right=10pt,
    top=8pt,
    bottom=8pt
]
{\ttfamily\fontsize{9pt}{10.5pt}\selectfont
\setlength{\parindent}{0pt}
\setlength{\parskip}{2pt}

You are an AI assistant specializing in point-level visual grounding for robot arm camera images.

Given the robot arm camera image and the task instruction: \{task\_text\}.

Generate diverse question-answer pairs focused on point-level visual grounding.

Question Types:

Type 1: FUNCTIONAL\_INTERACTION\_POINT.\\
Locate a point on a functional part of an object that is suitable for an action such as grasping, pushing, pulling, balancing, or pressing.\\
Example: Which point on the [object] should the robot gripper grasp to lift it securely?\\
Example: Which point on the [object] is the optimal place to apply pressure to activate it?

Type 2: PHYSICAL\_CAUSAL\_POINT.\\
Locate the contact point where the gripper would first touch an object, or, if the object serves as a container or target, the point where another item would land or rest.\\
Example: If the robot were to place the [object] down, which point would it first contact on the surface?\\
Example: Which point indicates where the [object] should be placed to complete the task?

Type 3: SPATIAL\_EXTREME\_POINT.\\
Locate a point representing a spatial extreme among visible objects, such as highest, lowest, closest, farthest, leftmost, or rightmost.\\
Example: Which point is on the part of the [object] that is furthest from the camera?\\
Example: Which point marks the highest location on the [object] visible in the scene?

Type 4: RELATIONAL\_BETWEEN\_POINT.\\
Locate a point on an object or surface that lies spatially between two other visible objects, or at the midpoint of a spatial relationship.\\
Example: Which point on the [object] is equidistant from [object A] and [object B]?

Type 5: SURFACE\_OR\_PART\_POINT.\\
Locate a point explicitly on a specific surface or named part of an object, such as a handle, edge, base, tip, rim, corner, or center.\\
Example: Which point is on the tip of the [object]'s [part]?\\
Example: Which point lies on the upper edge of the [object]?

Additional Rules:

1. Generate exactly 10 QA pairs: 2 questions for each type.\\
2. The questions should be diverse, covering different objects, viewpoints, and reasoning patterns. Do not simply copy the examples.\\
3. All questions must start with "Which point" or "If" while maintaining natural fluency and diverse expressions.\\
4. The answers should cover both single-point and multi-point cases. Some questions may return one coordinate, while others may return multiple coordinates when multiple contact points or symmetrical parts exist.\\
5. Interpret spatial relations, including up/down, front/back, left/right, and near/far, according to the 3D space of the image. Front/back relations refer to depth, not vertical position.\\
6. Each answer should contain only coordinates in the specified format. Do not include any reasoning process.

Output Format:

Return the result as a JSON array ONLY. Do not include any explanations or extra text:

[\\
\hspace*{1em}\{\\
\hspace*{2em}"Question": "Your point grounding question here",\\
\hspace*{2em}"Answer": [\\
\hspace*{3em}\{"point\_2d": [x, y]\}\\
\hspace*{2em}]\\
\hspace*{1em}\},\\
\hspace*{1em}\{\\
\hspace*{2em}"Question": "Your point grounding question here",\\
\hspace*{2em}"Answer": [\\
\hspace*{3em}\{"point\_2d": [x, y]\},\\
\hspace*{3em}\{"point\_2d": [x, y]\}\\
\hspace*{2em}]\\
\hspace*{1em}\}\\
]

Generate exactly 10 question-answer pairs.
}
\end{tcolorbox}


\begin{tcolorbox}[
    breakable,
    width=\linewidth,
    colback=orange!3,
    colframe=orange!75!black,
    coltitle=white,
    colbacktitle=orange!85!black,
    title=Prompt for Spatial Reasoning,
    fonttitle=\bfseries\large,
    boxrule=0.8pt,
    arc=3pt,
    left=10pt,
    right=10pt,
    top=8pt,
    bottom=8pt
]
{\ttfamily\fontsize{9pt}{10.5pt}\selectfont
\setlength{\parindent}{0pt}
\setlength{\parskip}{2pt}

You are an AI assistant specializing in spatial intelligence analysis of robot gripper camera images and task instructions.

Given the robot gripper camera image and the task instruction: \{task\_instruction\}.

The task instruction is human-annotated and may be incorrect, incomplete, or null; use it only as a reference.

Generate natural question-answer pairs focused on spatial intelligence.

Candidate Question Categories:

1. Object counting: How many [object A] are visible in the scene?\\
2. Spatial relationships: What is the relative position of [object A] compared to [object B]?\\
3. Distance estimation: How far apart are [object A] and [object B]?\\
4. Spatial orientation: In which direction is [object A] oriented relative to the robot gripper?\\
5. Depth perception: Which object appears closer or farther from the camera?\\
6. Geometric properties: What is the approximate shape or size of [object A] in the scene?\\
7. Spatial arrangement: How are the objects arranged in the workspace?\\
8. Accessibility analysis: Which objects are within the robot's reach based on their positions?\\
9. Collision avoidance: What spatial constraints must the robot consider when moving toward [object A]?\\
10. 3D spatial understanding: What is the vertical or horizontal relationship between objects?

Additional Rules:

1. Select 4 categories from the candidate categories and generate exactly 2 questions for each selected category, for a total of 8 QA pairs.\\
2. Each question must require image-based spatial information and manipulation-relevant physical reasoning, such as gravity, friction, stability, grasp safety, reachability, or collision constraints. Questions answerable by external knowledge alone are prohibited.\\
3. Questions should be contextually grounded, colloquial, and diverse in category, object combination, reasoning angle, and manipulation aspect, such as motion planning, safety, feasibility, manipulation difficulty, or environmental interaction.\\
4. Questions should include only minimal explicit information and require the model to infer relevant details from the image. Use multiple distinct objects when possible, and avoid isolated focus on a single object.\\
5. Answer only the specific question asked, without task background, general advice, future steps, unrelated spatial observations, or expansion phrases such as "additionally", "furthermore", or "also note".\\
6. Keep each answer scoped to the requested attribute: if the question asks about distance, answer only about distance; if it asks about orientation, answer only about orientation.\\
7. Interpret front, back, left, right, up, and down according to the 3D space of the image.

Quantity Constraint [STRICT]:

Exactly 2 questions per selected category.\\
Total: 8 question-answer pairs.\\
No more, no less.

Length Constraint [STRICT]:

Question: \textless{} 50 words.\\
Answer: \textless{} 150 words.

Output Format:

Return the result as a JSON array ONLY. Do not include any explanations or extra text:

[\\
\hspace*{1em}\{\\
\hspace*{2em}"Question": "Your spatial intelligence question here",\\
\hspace*{2em}"Answer": "Your detailed spatial analysis answer here - STRICTLY LIMITED to question scope"\\
\hspace*{1em}\},\\
\hspace*{1em}\{\\
\hspace*{2em}"Question": "Your spatial intelligence question here",\\
\hspace*{2em}"Answer": "Your detailed spatial analysis answer here - STRICTLY LIMITED to question scope"\\
\hspace*{1em}\}\\
]

Repeat for all 8 pairs.
}
\end{tcolorbox}

\subsection{RefSpatial-based Fisheye VQA Construction}
\label{app:refspatial_fisheye_vqa}

For the spatial-diversity supplement, we start from RefSpatial images and their associated VQA annotations. Since the original images are captured from standard perspectives, we first convert them into fisheye-style images using FLUX.2-dev. The editing prompt asks the model to preserve the original scene content while applying fisheye-style radial distortion, so that the resulting images remain semantically consistent with the original RefSpatial scenes while better matching the wrist-fisheye visual regime. We then filter the original RefSpatial VQA annotations before adding them to UMI-VQA. In particular, we discard QA pairs whose answers contain explicit 2D pixel coordinates, because fisheye conversion changes pixel locations even when the scene content and viewing direction are approximately preserved. Keeping such coordinate-based annotations would make the answer inconsistent with the transformed image. Non-coordinate QA pairs are retained as the RefSpatial-based spatial-diversity supplement.


\begin{tcolorbox}[
    breakable,
    width=\linewidth,
    colback=orange!3,
    colframe=orange!75!black,
    coltitle=white,
    colbacktitle=orange!85!black,
    title=FLUX.2 Fisheye-transform Prompt,
    fonttitle=\bfseries\large,
    boxrule=0.8pt,
    arc=3pt,
    left=10pt,
    right=10pt,
    top=8pt,
    bottom=8pt
]
{\ttfamily\fontsize{9pt}{10.5pt}\selectfont
\setlength{\parindent}{0pt}
\setlength{\parskip}{3pt}

Apply a fisheye-style transformation to the input image.

Strict requirements:

1. Preserve the original scene semantics, object identities, and relative spatial relations, especially in the central region of the image. \\
2. Expand the field of view to create a wide-angle fisheye-style view of the same scene, with an approximate fov of 150$^\circ$. \\
3. Keep the original camera direction and central viewing angle unchanged. Do not rotate the camera or alter the main viewpoint. \\
4. Apply fisheye-style radial distortion: distortion should be minimal near the image center and progressively stronger toward the edges. \\
5. When expanding beyond the original frame, generate only plausible peripheral scene content consistent with the original image. \\
6. Preserve the original lighting, scene structure, and overall composition as much as possible. \\
7. The result should look like the same scene captured by a fisheye lens with a wider field of view, not a different or newly imagined scene. \\
8. Do not add unrelated objects, remove key scene elements, or introduce stylized effects such as tiny-planet, spherical-world, or mirror-ball rendering. \\
}
\end{tcolorbox}

\section{Implementation Details and Experimental Results}
\label{app:more_exp_details}

\paragraph{RoboTwin and LIBERO wrist-view fisheye data generation.}
We generate the wrist-view fisheye data with a two-stage pipeline. For RoboTwin,
we first collect demonstrations using wide-angle wrist cameras whose vertical
field of view is $150^\circ$ and whose rendered resolution is $680 \times 680$.
For the standard collection configuration, each task is collected with 50
successful episodes. For LIBERO, we use the same image-generation and
post-processing pipeline, but remove the third-person view and keep only a
single wrist camera. The retained LIBERO wrist camera uses a rendered resolution
of $700 \times 700$, while sharing the same fisheye distortion parameters as
RoboTwin.

After collecting the raw episodes, we convert the raw HDF5 files into LeRobot
format and apply a fisheye-style image transform during this conversion. For
each wrist image, we build a fixed remapping grid based on the image size. Let
$(c_x,c_y)$ be the image center and $R=\min(W,H)/2$. For an output pixel
$(u,v)$, we define the normalized radius
\[
\rho =
\sqrt{\left(\frac{u-c_x}{R}\right)^2 +
      \left(\frac{v-c_y}{R}\right)^2 } .
\]
With fisheye strength $s=1.8$, the corresponding source radius is
\[
\rho_{\mathrm{src}} =
\frac{\tan\left(\rho \arctan(s)\right)}{s}.
\]
The source coordinate is then obtained by scaling the normalized ray from the
image center by $\rho_{\mathrm{src}} / \rho$. Pixels outside the unit circle
($\rho > 1$) are mapped outside the image and filled with a constant black
border. We apply the remapping with bilinear interpolation using OpenCV
\texttt{remap}, and then resize the result to $224 \times 224$ with area
interpolation.

\paragraph{Additional Experimental Details.} During evaluation, VISTA and all baselines are trained and evaluated using the same datasets and the same number of training steps. For the simulation experiments on LIBERO and RoboTwin, we perform multi-task mixed training for 40k steps with a batch size of 32. On LIBERO, each task is evaluated over 50 trials, while on RoboTwin, each task is evaluated over 100 trials. For the real-world experiments, we perform single-task fine-tuning for 20k steps on each task, and report the success rate over 20 trials. 
The detailed experimental results are shown in Table~\ref{tab:robotwin_result_all}, Table \ref{tab:libero_result_all} and Table \ref{tab:real_results_all}

\begin{table}[t]
\centering
\small
\setlength{\tabcolsep}{4pt}
\begin{tabular}{lcccc}
\toprule
Task & LingBot-VLA & $\pi_{0.5}$ & Wall-X & VISTA \\
\midrule
beat\_block\_hammer & 87.0 & 95.0 & 33.0 & \textbf{99.0} \\
click\_bell & 12.0 & 81.0 & 41.0 & \textbf{91.0} \\
grab\_roller & 78.0 & 77.0 & 28.0 & \textbf{84.0} \\
lift\_pot & 75.0 & 73.0 & 15.0 & \textbf{91.0} \\
move\_can\_pot & 64.0 & 68.0 & 14.0 & \textbf{71.0} \\
move\_playingcard\_away & 34.0 & 50.0 & 4.0 & \textbf{58.0} \\
open\_microwave & 23.0 & 40.0 & 15.0 & \textbf{62.0} \\
pick\_dual\_bottles & 48.0 & 64.0 & 8.0 & \textbf{84.0} \\
pick\_diverse\_bottles & 55.0 & 34.0 & 2.0 & \textbf{65.0} \\
place\_bread\_basket & 51.0 & 54.0 & 3.0 & \textbf{72.0} \\
place\_bread\_skillet & 60.0 & 61.0 & 5.0 & \textbf{67.0} \\
place\_can\_basket & 58.0 & 61.0 & 8.0 & \textbf{65.0} \\
place\_mouse\_pad & 13.0 & 14.0 & 8.0 & \textbf{25.0} \\
place\_object\_scale & 9.0 & 29.0 & 1.0 & \textbf{31.0} \\
place\_phone\_stand & 45.0 & 71.0 & 2.0 & \textbf{73.0} \\
press\_stapler & 69.0 & 69.0 & 31.0 & \textbf{71.0} \\
rotate\_qrcode & 44.0 & 56.0 & 14.0 & \textbf{58.0} \\
stack\_bowls\_two & 90.0 & 90.0 & 25.0 & \textbf{92.0} \\
stamp\_seal & 39.0 & 53.0 & 43.0 & \textbf{57.0} \\
turn\_switch & 44.0 & 48.0 & 4.0 & \textbf{49.0} \\
\midrule
Average & 49.9 & 59.4 & 15.2 & \textbf{68.3} \\
\bottomrule
\end{tabular}
\caption{Per-task success rates (\%) on the umi-style RoboTwin benchmark.}
\label{tab:robotwin_result_all}
\end{table}

\begin{table}[t]
\centering
\small
\setlength{\tabcolsep}{5pt}
\begin{tabular}{lccccc}
\toprule
Policy & LIBERO-10 & Goal & Spatial & Object & Average \\
\midrule
$\pi$-0.5 & 87.0 & 87.6 & 95.4 & \textbf{98.8} & 92.2 \\
LingBot-VLA & 65.6 & 77.8 & 88.2 & 95.2 & 81.7 \\
Wall-X & 54.6 & 63.0 & 79.0 & 83.2 & 70.0 \\
VISTA & \textbf{88.8} & \textbf{91.6} & \textbf{97.8} & \textbf{98.8} & \textbf{94.3} \\
\bottomrule
\end{tabular}
\caption{Success rates (\%) on umi-style LIBERO benchmark suites.}
\label{tab:libero_result_all}
\end{table}

\begin{table*}[t]
\centering
\small
\setlength{\tabcolsep}{4pt}
\begin{tabular}{lccc}
\toprule
\textbf{Task} & \textbf{LingBot-VLA} & \textbf{$\pi_{0.5}$} & \textbf{VISTA} \\
\midrule
Close Laptop and Place Mouse        & 0.20 & 0.30 & \textbf{0.55} \\
Place Dolls into Box                & 0.45 & 0.25 & \textbf{0.55} \\
Take Dolls out of Box               & 0.30 & \textbf{0.65} & 0.55 \\
Place Stapler on Cabinet            & 0.65 & \textbf{0.85} & \textbf{0.85} \\
Stack Side Cubes on Center Cube     & 0.00 & 0.45 & \textbf{0.50} \\
Sort Cubes by Color into Tray       & 0.35 & 0.40 & \textbf{0.55} \\
Retrieve Toast from Toaster         & 0.55 & 0.35 & \textbf{0.70} \\
Pick Target Fruits from Bowl        & \textbf{0.35} & \textbf{0.35} & 0.25 \\
Put Doll into Drawer and Close      & 0.45 & \textbf{0.90} & 0.80 \\
Open Drawer                         & 0.00 & 0.40 & \textbf{0.55} \\
Organize Dolls                      & 0.55 & \textbf{0.80} & \textbf{0.80} \\
Place Bun into Rice Cooker and Close & 0.35 & 0.60 & \textbf{0.65} \\
Arrange Flowers                     & 0.55 & \textbf{0.75} & 0.55 \\
Place Drink into Box                & 0.00 & 0.25 & \textbf{0.40} \\
Hang Mug on Rack                    & 0.00 & \textbf{0.60} & 0.50 \\
Stack Pen Holders                   & 0.00 & 0.40 & \textbf{0.55} \\
Pour Chips from Bowl to Plate       & 0.80 & \textbf{0.85} & 0.70 \\
Pick Plum from Cluttered Fruits     & 0.00 & 0.65 & \textbf{0.85} \\
Stack Paper Cups                    & 0.10 & 0.35 & \textbf{0.45} \\
Place Fruits                        & 0.60 & 0.40 & \textbf{0.65} \\
\midrule
\textbf{Overall}                    & 0.313 & 0.528 & \textbf{0.598} \\
\bottomrule
\end{tabular}
\caption{
Real-robot evaluation on 20 UMI-collected manipulation tasks.
All methods are trained on the same validated UMI dataset and evaluated with 20 trials per task.
Success rates are reported in $[0,1]$.
LingBot-VLA tasks without a valid executable result are counted as zero.
}
\label{tab:real_results_all}
\end{table*}

\end{document}